\documentclass[10pt,reqno]{amsart}

\usepackage{companypaper}
\usepackage{algorithm}
\usepackage{algorithmic}
\usepackage{natbib}
\usepackage{hyphenat}
\usepackage{enumitem}
\usepackage{verbatim}

\makeatletter
\pretocmd{\@settitle}{\let\uppercasenonmath\@gobble}{}{}
\patchcmd{\@settitle}{\bfseries}{\bfseries\LARGE}{}{}
\pretocmd{\@setauthors}{\let\MakeUppercase\@firstofone}{}{}
\patchcmd{\@setauthors}{\centering\footnotesize}{\centering\large}{}{}
\apptocmd{\@setauthors}{}{}{}
\renewcommand{\@setaddresses}{}
\renewcommand{\@setdate}{%
  \noindent\normalfont\footnotesize
  \textsuperscript{1}Fujitsu Limited,~ 
  \textsuperscript{2}RIKEN Center for AIP,~
  \textsuperscript{3}Institute of Science Tokyo,~
  \textsuperscript{4}Tokai University,~
  \textsuperscript{5}The University of Tokyo\par
  \vskip 2pt
  \textbf{Correspondence}: Yuma Ichikawa at \textcolor{blue}{\texttt{ichikawa.yuma@fujitsu.com}}\par
}
\makeatother

\usepackage{pifont}
\usepackage{booktabs}
\usepackage{multirow}
\usepackage{threeparttable}
\usepackage{array}

\usepackage{fontawesome5}
\newcommand{\ghurl}[2][]{%
  \href{#2}{\raisebox{-0.1em}{\faGithub}\,#1\nolinkurl{#2}}%
}

\newtcolorbox{codebox}[1][]{
  sharp corners,
  colback=PyCodeBg,
  colframe=PyCodeBg,
  boxrule=0pt,
  left=0mm, right=0mm, top=1mm, bottom=1mm,
  #1
}

\usepackage{newfloat}
\DeclareFloatingEnvironment[
  fileext=loc,
  listname={List of Codes},
  name=Code,
  placement=t,
]{codefloat}

\usepackage{nicefrac}
\usepackage{physics2}
\usepackage{fixdif, derivative}
\usephysicsmodule{ab}

\newcommand{\mac}[1]{{\mathcal #1}}

\DeclareMathOperator*{\argmin}{arg min}

\newcommand{\onecomp}{\textsc{OneComp}\xspace}
\newcommand{\autobit}{AutoBit\xspace}

\newcommand{\qep}{QEP\xspace}
\newcommand{\lpcd}{LPCD\xspace}
\newcommand{\jointq}{JointQ\xspace}
\newcommand{\mdbf}{MDBF\xspace}

\CompanyLogo{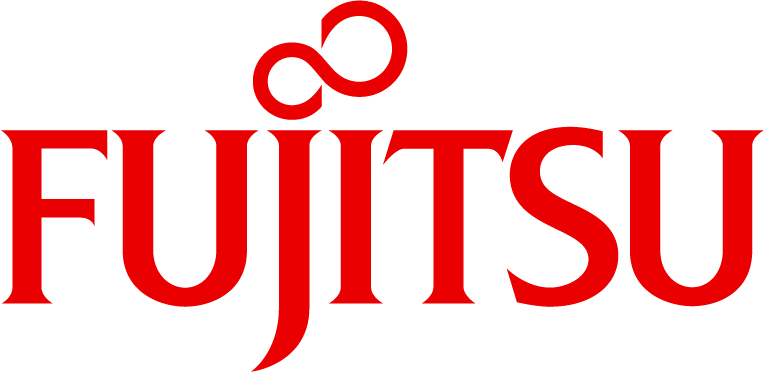}
\CompanyLogoHeight{11mm}
\CompanyHeaderText{Fujitsu Limited}
\CompanyDocType{}
\CompanyClassification{}
\CompanySubtitle{}
\CompanyRunningTitle{OneComp: One-Line Revolution for Generative AI Model Compression}

\title{OneComp: One-Line Revolution for \\ Generative AI Model Compression}

\author{
Yuma Ichikawa\textsuperscript{1,2},
Keiji Kimura\textsuperscript{1},
Akihiro Yoshida\textsuperscript{1,3},
Yudai Fujimoto\textsuperscript{1,3},
Hiroki Tokura\textsuperscript{1},\\[2pt]
Yamato Arai\textsuperscript{1,5},
Yoshiyuki Ishii\textsuperscript{1},
Yusei Kawakami\textsuperscript{1},
Genki Shikada\textsuperscript{1},
Achille Jacquemond\textsuperscript{1},\\[2pt]
Yoshihiko Fujisawa\textsuperscript{1,3},
Katsuki Fujisawa\textsuperscript{3},
Takumi Honda\textsuperscript{1},
Akira Sakai\textsuperscript{1,4}}
\date{March 31, 2026}

\begin{document}

\maketitle
\begin{center}
\small
\ghurl{https://github.com/FujitsuResearch/OneCompression}
\end{center}

\begin{abstract}
Deploying foundation models is increasingly constrained by memory footprint, latency, and hardware costs. Post-training compression can mitigate these bottlenecks by reducing the precision of model parameters without significantly degrading performance; however, its practical implementation remains challenging as practitioners navigate a fragmented landscape of quantization algorithms, precision budgets, data-driven calibration strategies, and hardware-dependent execution regimes. We present OneComp, an open-source compression framework that transforms this expert workflow into a reproducible, resource-adaptive pipeline. Given a model identifier and available hardware, OneComp automatically inspects the model, plans mixed-precision assignments, and executes progressive quantization stages, ranging from layer-wise compression to block-wise refinement and global refinement. A key architectural choice is treating the first quantized checkpoint as a deployable pivot, ensuring that each subsequent stage improves the same model and that quality increases as more compute is invested.
By converting state-of-the-art compression research into an extensible, open-source, hardware-aware pipeline, OneComp bridges the gap between algorithmic innovation and production-grade model deployment.
\end{abstract}

\section{Introduction}

\begin{figure}[t]
  \centering
  \includegraphics[width=\linewidth]{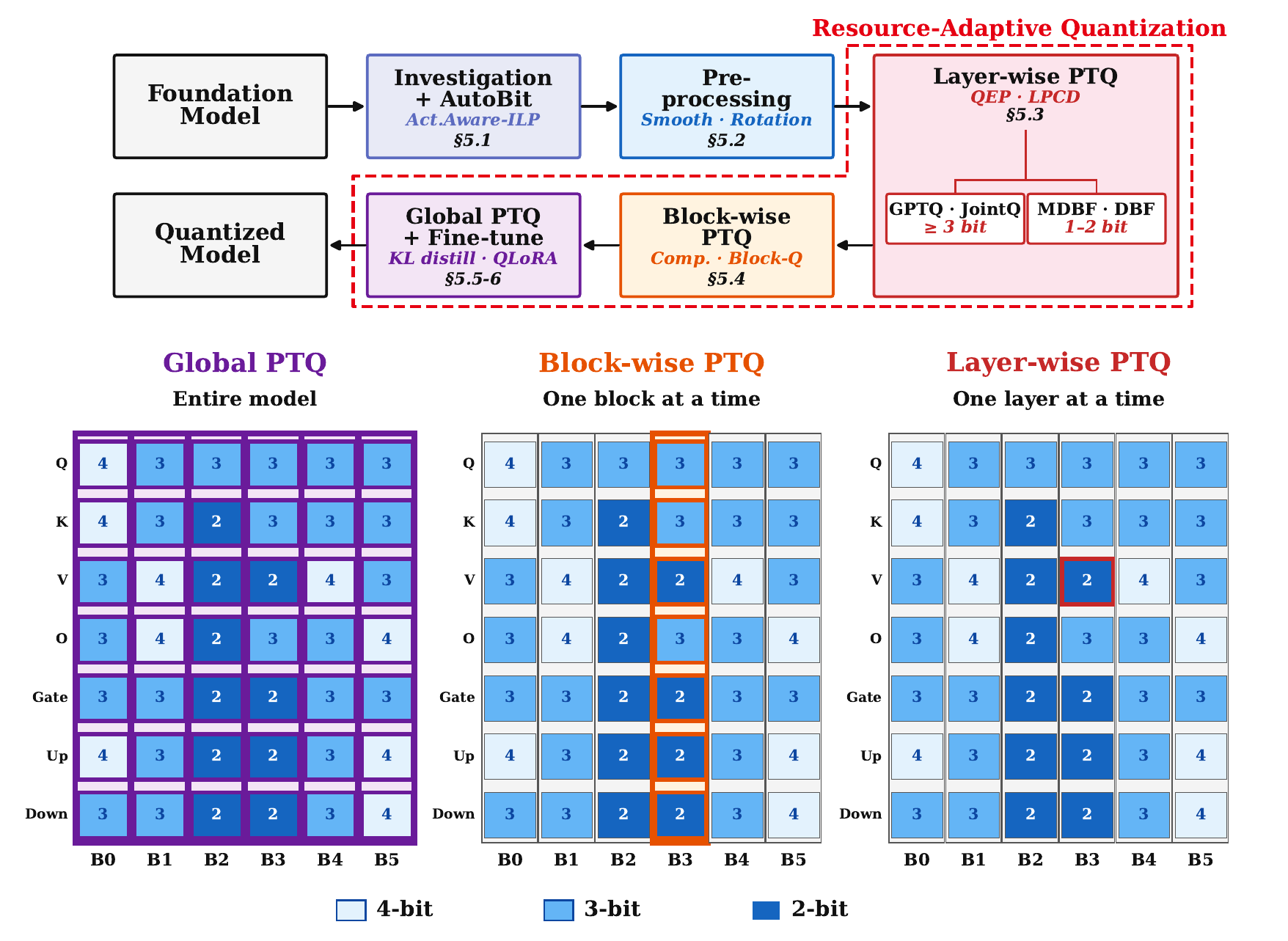}
  \caption{Overview of the \onecomp pipeline.
  \emph{Top}: end-to-end workflow from a pretrained foundation model to a deployment-ready quantized model.
  \emph{Bottom}: the three quantization granularity levels on a Transformer with mixed-precision bit allocation.
  Cell colors indicate per-layer bit-widths assigned by \autobit: Darker colors indicate lower precision.
  Layer-wise PTQ operates on one linear layer; block-wise PTQ on one Transformer block; global PTQ on the full model.}
  \label{fig:workflow}
\end{figure}

Foundation models have become the backbone of modern generative AI, demonstrating strong performance in reasoning, coding, and other tasks \citep{touvron2023llama, touvron2023llama2, achiam2023gpt, liu2024deepseekv3, team2024gemini}; however, this rapid growth poses significant challenges for practical deployment due to substantial memory footprints and computing requirements.
Models with tens to hundreds of billions of parameters greatly exceed the memory capacity of commodity hardware, making it prohibitive for most organizations to serve them at full precision.
Consequently, model quantization has become a key strategy to bridge the gap between massive model sizes and limited computational resources.
Unlike Quantization-Aware Training (QAT) which requires extensive retraining, Post-Training Quantization (PTQ) addresses these challenges by reducing the precision of weights and activations after training, thereby directly reducing memory footprint and accelerating inference~\citep{gong2024survey,gholami2022survey,nagel2021white}.

The research community has responded with a rich yet fragmented ecosystem of compression methods.
Hessian-aware rounding~\citep{frantar2022gptq}, activation-aware scaling~\citep{lin2024awq}, rotation-based outlier suppression~\citep{tseng2024quip,ashkboos2024quarot,liu2024spinquant}, block-wise joint optimization~\citep{li2021brecq,ding2023cbq,shao2023omniquant}, and structured binary-factor formats~\citep{xu2024onebit,boza2026addition,ichikawa2025mdbf} address different aspects of the problem, with varying assumptions, hyperparameters, and failure modes.
Moreover, the optimal combination of algorithms, compression ratios, and configurations varies across models, tasks, and hardware~\citep{pmlr-v139-hubara21a,zhao2025benchmarking}, with no single recipe applicable universally.
Consequently, the latest advances in compression research often take considerable time to reach practitioners, creating a significant gap between what is theoretically achievable and what is practically deployed.
What is needed is not another algorithm but a principled end-to-end workflow that manages complexity related to algorithm selection, bit-width planning, and multi-stage sequencing on behalf of the user.
In many areas of machine learning, the shift from manual configuration to automated pipelines has facilitated broad adoption~\citep{conti2020technical, he2021automl}.
It is time for model quantization to follow a similar paradigm shift.

\onecomp is the open-source compression package that facilitates this transition.
Given a model identifier and available hardware, \onecomp automatically constructs a complete quantization workflow.
The pipeline proceeds through the following stages, illustrated in Figure~\ref{fig:workflow}.
First, an \emph{investigation phase} profiles the quantization sensitivity of each layer and solves a constrained optimization problem to plan heterogeneous bit-widths across the model (Section~\ref{sec:autobit}).
Next, optional \emph{preprocessing} transforms redistribute outlier energy to enhance quantization friendliness (Section~\ref{sec:preprocess}).
The core \emph{quantization cascade} then executes in a resource-adaptive manner.
When limited GPU memory is available, \onecomp applies layer-wise PTQ (Section~\ref{sec:layerwise}), processing one linear layer at a time; this mode is enhanced by closed-form error-propagation corrections~\citep{arai2025quantization} and submodule-aware coordinate descent~\citep{ichikawa2025lpcd}, allowing it to achieve quality comparable to more expensive methods.
When more resources are available, the pipeline seamlessly integrates block-wise PTQ (Section~\ref{sec:blockwise}), including block-wise distillation and model-level KL-divergence optimization, with each stage improving upon the last.
A final optional fine-tuning stage via QLoRA~\citep{dettmers2023qlora} enables the global PTQ to recover any remaining gap (Section~\ref{sec:global} and~\ref{sec:finetuning}).
The user always receives the best quantized model achievable within the given hardware budget, with no manual configuration required.

The design is based on three principles.
First, a \textbf{path-to-plan API} dynamically derives the workflow from the model architecture and GPU budget, freeing users from manual design.
Second, a \textbf{resource-adaptive engine} provides monotonically improving quality, ranging from layer-wise PTQ through block-wise refinement to global PTQ and fine-tuning.
Third, an \textbf{extensible refiner architecture} ensures that new algorithms can be integrated without the need to redesign the pipeline.
Based on these principles, \onecomp covers the full bit-width spectrum.
At 3--4 bits, a joint scale-and-integer optimizer substantially outperforms standard GPTQ while producing deployment-compatible checkpoints.
At 1--2 bits, structured binary-factor formats maintain meaningful accuracy where uniform quantization fails.
Furthermore, an automatic mixed-precision planner allocates heterogeneous bit-widths within a memory budget.
Experimental evaluation across the LLaMA~\citep{touvron2023llama,grattafiori2024llama} and Qwen~\citep{yang2025qwen3} families demonstrates consistent improvements over existing tools and monotonic quality gains as more resources are invested.

\section{Background}
\label{sec:background}

This section introduces the background needed to understand quantization in foundation models. 
We begin by motivating quantization in this context, emphasizing the memory-capacity and bandwidth bottlenecks that affect large-scale inference (Section~\ref{sec:why_quantization_matters}). 
We then describe the historical context of quantization (Section~\ref{subsec:signel2nn}).
Next, we review the standard formulation of uniform quantization and identify the quantization targets in Transformer architectures (Section~\ref{sec:quant_fundamentals} and~\ref{sec:what_quantize}).
After that, we introduce post-training quantization (PTQ) and distinguish it from quantization-aware training (QAT) (Section~\ref{sec:ptq_objective}). 
We conclude by describing three practical PTQ regimes: layer-wise, block-wise, and global, which are central to designing \onecomp (Section~\ref{sec:three_levels}). Although current interest in LLM quantization is largely driven by deployment needs, the underlying concepts trace back to the signal-processing literature, which we briefly revisit at the end of this section.

\subsection{Why Quantization Matters for Foundation Models}
\label{sec:why_quantization_matters}

Quantization reduces the number of bits used to represent numerical values, thereby storing each value more compactly. Replacing a 16-bit representation with 4 bits reduces storage by a factor of four; using 2 bits yields an eightfold reduction. For small models, this can seem to be a convenient compression technique. However, for foundation models, numerical precision is a first-order systems concern: it directly determines whether the model fits in memory, the amount of data transferred during inference, and deployment feasibility.

The most immediate pressure comes from model weights. Modern Transformer-based foundation models typically contain billions to tens of billions of parameters, with the majority residing in large linear projection matrices. For a model with $P$ parameters, storing weights in 16-bit precision requires $2P$ bytes of memory. A 70-billion-parameter model requires approximately 140 GB to store its weights in 16-bit precision, compared to about 35 GB at 4 bits and roughly 18 GB at 2 bits. These estimates overlook the minor overhead associated with storing scales and zero-points while capturing the main savings. In practice, such reductions often determine whether a model can be deployed on a single accelerator or must be distributed across multiple devices.
Memory capacity, however, is only part of the picture. Inference in Transformer models is often limited more by memory bandwidth than by raw arithmetic throughput~\citep{yuan2024llm}. At each inference step, the accelerator fetches large weight matrices from high-bandwidth memory for matrix multiplications. Reducing weight precision decreases the volume of data transferred for the same computation, which can improve throughput, reduce latency, and lower energy consumption~\citep{gong2024survey,dettmers2023case}. Quantization is valuable not only for compressing the model in memory but also for reducing data movement costs across the system.

In addition to static weights, foundation-model inference involves large dynamic tensors. The most important of these are the intermediate activations and the key--value (KV) cache used in autoregressive decoding. Activations are the intermediate tensors produced as input tokens propagate through the network. The KV cache stores the key and value tensors from previous tokens, allowing the model to avoid recomputing attention over the previously generated tokens at each decoding step. As context length or batch size increases, the memory required for these tensors can become substantial, potentially approaching the memory consumed by the weights themselves~\citep{ashkboos2024quarot}. For this reason, modern quantization methods increasingly target not only weights but also activations and KV caches; Section~\ref{sec:what_quantize} revisits these components in detail.

\onecomp focuses on weight-only PTQ, providing a compelling entry point for compressing foundation models. Since most parameters are stored in the weights, quantizing them results in a significant and immediate reduction in memory footprint. Moreover, weights can be quantized offline after training without altering the original optimization pipeline and can be executed using specialized low-precision kernels during inference. The subsequent subsections discuss activation and KV-cache quantization for completeness; however, the core method in \onecomp emphasizes weight quantization.

\subsection{Historical Context}
\label{subsec:signel2nn}

Although the deployment setting above is specific to modern foundation models, the concept of quantization predates deep learning. Early work in classical signal processing explored how to map continuous-valued signals to a finite set of reconstruction levels while minimizing distortion under a fixed bit budget. In particular, the seminal analyses of \citet{max1960quantizing} and \citet{lloyd1982least}, widely known as Lloyd--Max quantization, showed that optimal quantization levels should adapt to the underlying signal distribution, assigning finer resolution to regions with higher probability mass. These results established the mathematical foundation for later developments in rate--distortion theory and continue to inform modern quantization methods.

Quantization entered neural-network research as model size, memory traffic, and deployment costs became increasingly important in deep learning. Early studies demonstrated that many neural networks remain effective with reduced numerical precision. For example, \citet{gupta2015deeplearning} showed that low-precision arithmetic can support both training and inference across various workloads, while \citet{courbariaux2015binaryconnect} and \citet{courbariaux2016binarynet} extended on this concept using binary weights and activations. Subsequent work, including \citet{zhou2016dorefa} and \citet{hubara2018quantized}, expanded on these ideas to low-bit training regimes.
Meanwhile, research on model compression established quantization as a practical systems technique rather than just a numerical abstraction. \citet{han2016deep} showed that pruning, trained quantization, and entropy coding can compress neural networks by $35$--$49\times$ with negligible loss in accuracy, highlighting the substantial redundancy in overparameterized models. A major turning point came with \citet{jacob2018quantization}, which popularized the deployment-oriented paradigm of integer-only inference, where weights and activations are quantized to enable efficient low-precision arithmetic while largely retaining floating-point accuracy. This research helped establish quantization as a tool not only for compression but also for reducing latency, memory bandwidth demand, and energy consumption on real hardware.

Today, quantization is one of the most widely used techniques for compressing and accelerating neural networks, particularly in deployment scenarios where data movement can be as costly as computation itself~\citep{nagel2021white,gholami2022survey,gong2024survey}. The methods examined in this paper belong to this modern line of work: they inherit the basic formalism of classical quantization but are designed to accommodate the scale and system constraints of contemporary Transformer-based foundation models.

\subsection{Fundamentals of Uniform Quantization}
\label{sec:quant_fundamentals}

Uniform quantization approximates real numbers using a finite grid of evenly spaced levels. Two central questions arise: how a real value maps onto this grid and how broadly the corresponding quantization parameters are shared. We review both below.

\subsubsection{Quantization Mapping}

\begin{figure}[tb]
    \centering
    \includegraphics[width=\linewidth]{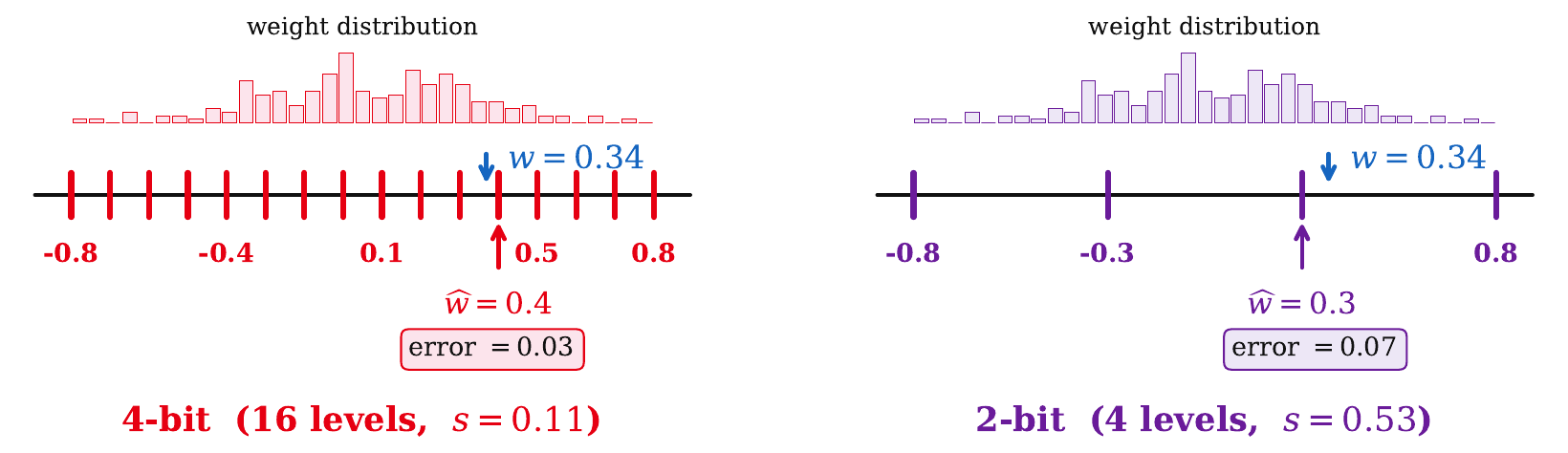}
    \caption{Comparison of 4-bit and 2-bit uniform quantization on the same weight distribution.
    Each panel overlays the quantization grid on a histogram of weights.
    4-bit uniform quantization provides 16 fine-grained levels; 2-bit uniform quantization provides only 4, yielding coarser approximation and larger errors.}
    \label{fig:quant_illustration}
\end{figure}

Quantization replaces a continuous value with one of a finite number of representable levels. The most common scheme is uniform quantization, in which adjacent levels are evenly spaced. For a scalar value $w \in \mathbb{R}$ and a target bit-width $b$, a uniform quantizer is parameterized by a scale $s > 0$ and a zero-point $z \in \mathbb{Z}$. Let $q_{\min}$ and $q_{\max}$ denote the available integer range; for unsigned $b$-bit quantization, $q_{\min}=0$ and $q_{\max}=2^b-1$. The quantization and dequantization operations are
\begin{equation}
  q = \mathrm{clamp}\left(\left\lfloor \frac{w}{s} \right\rceil + z,\, q_{\min},\, q_{\max}\right), \qquad
  \widehat{w} = s \cdot (q - z),
  \label{eq:scalar_quant}
\end{equation}
where $\lfloor \cdot \rceil$ denotes rounding to the nearest integer and $\mathrm{clamp}(x,a,c)=\max(a,\min(x,c))$. The scale intuitively determines the spacing between adjacent quantization levels, while the zero-point aligns the real axis with the integer grid.
The integer $q$ is the low-precision value stored in memory, and the dequantized value $\widehat{w}$ is the real-valued approximation used in computation. The difference $w-\widehat{w}$ represents the quantization error. When $w$ is within the representable range, meaning that no clipping occurs, the rounding error is bounded by $|w-\widehat{w}| \le \nicefrac{s}{2}$. If $w$ falls outside the range $[s(q_{\min}-z),\, s(q_{\max}-z)]$, the clamp operation introduces additional clipping errors that can exceed \nicefrac{s}{2}. Figure~\ref{fig:quant_illustration} compares 4-bit and 2-bit quantization of the same weight distribution, illustrating how lower bit-widths create a coarser grid and consequently larger approximation errors.

\subsubsection{Choosing Scale and Zero-Point}

The scale and zero-point are usually calibrated from the observed range of values. Given a set of real numbers with a minimum $w_{\min}$ and a maximum $w_{\max}$, standard asymmetric min-max calibration~\citep{jacob2018quantization,czako2025activation_outlier,gholami2022survey} sets
\begin{equation}
  s = \frac{w_{\max} - w_{\min}}{q_{\max} - q_{\min}}, \qquad
  z = \mathrm{clamp}\!\left(\mathrm{round}\!\left(q_{\min} - \frac{w_{\min}}{s}\right),\, q_{\min},\, q_{\max}\right).
  \label{eq:scale_zp}
\end{equation}
This choice roughly maps the observed real range to the available integer range. It is simple, widely used, and particularly natural when the signal is not centered at zero.

When the distribution is approximately symmetric around zero, symmetric quantization is often preferred, particularly for weights. In this case, the zero-point is fixed at $z=0$, and the integer range is considered signed, typically $q \in \{-(2^{b-1}-1), \ldots, 2^{b-1}-1\}$. The scale is then selected as
\begin{equation}
  s = \frac{\max(|w_{\min}|, |w_{\max}|)}{2^{b-1}-1}, \qquad
  \widehat{w} = s \cdot q.
  \label{eq:sym_quant}
\end{equation}
Compared to asymmetric quantization, symmetric quantization simplifies dequantization into a single multiplication and eliminates the need to store a nonzero zero-point for each quantization group. This reduces metadata overhead and can simplify inference kernels~\citep{gholami2022survey}. In practice, asymmetric quantization is common for activations, while symmetric quantization is particularly prevalent for weights.

\subsubsection{Quantization Granularity}
\label{subsec:granularity}

\begin{figure}[t]
    \centering
    \includegraphics[width=\linewidth]{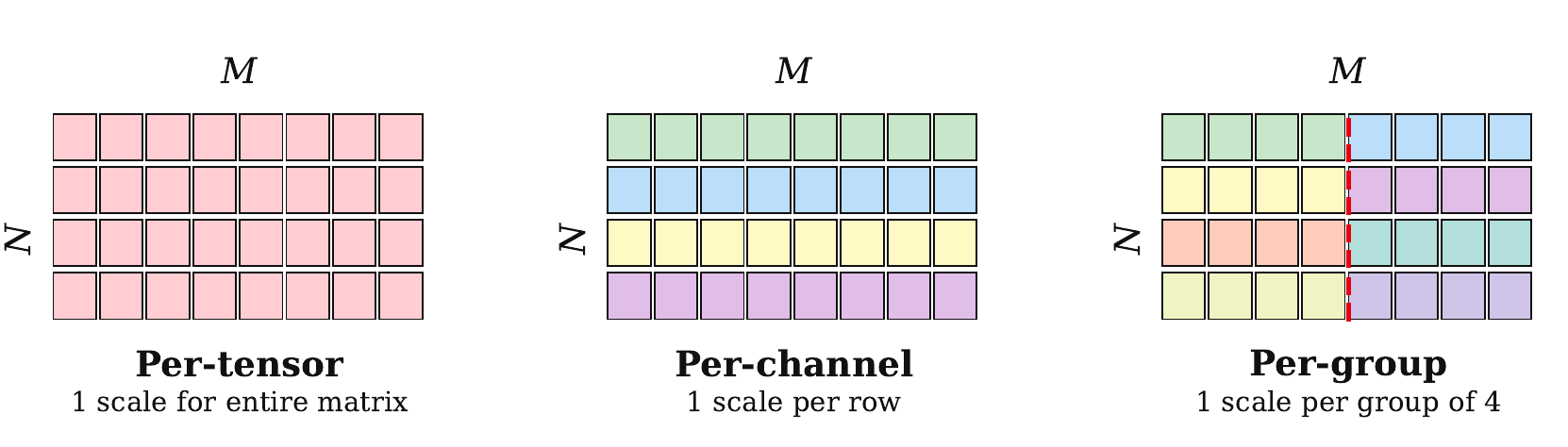}
    \caption{Three levels of quantization granularity for a weight matrix.
    Each color represents a group of elements sharing the same scale and zero-point.
    Per-tensor uses one set of parameters for the entire matrix; per-channel uses one per row; per-group divides each row into smaller groups.}
    \label{fig:granularity}
\end{figure}

Beyond bit-width, an important design choice is quantization granularity, that is, how many values share the same scale and zero-point. Finer granularity typically improves accuracy by better adapting to local variations in the data, but it also increases the number of auxiliary quantization parameters that must be stored. Figure~\ref{fig:granularity} illustrates three common granularities for weight matrices.

The simplest option is per-tensor quantization~\citep{gholami2022survey}, which uses a single scale and zero-point for the entire weight matrix. This is computationally convenient but can be inaccurate when different rows or columns have markedly different distributions. Per-channel quantization assigns separate parameters to each row or column, thereby capturing variation at the channel level. Per-group quantization, which is predominant in modern LLM quantization~\citep{frantar2022gptq}, offers a compromise between the two: each row is divided into groups of $G$ consecutive elements, and all elements within a group share the same quantization parameters. For a weight matrix $W \in \mathbb{R}^{N \times M}$, this can be written as
\begin{equation}
  \widehat{W}_{ij} = s_{g(i,j)} \left(q_{ij} - z_{g(i,j)}\right), \qquad q_{ij} \in \{0,1,\ldots,2^b-1\},
  \label{eq:uniform_quant}
\end{equation}
where $g(i,j)$ assigns each element to a quantization group.

A common choice in practice is $G=128$. Reducing the group size allows the quantizer to better adapt to local weight statistics, typically improving reconstruction quality. The trade-off is that smaller groups require more stored scales and zero-points. The total memory footprint of a per-group quantized matrix is approximately
\begin{equation}
  M_{\mathrm{group}} \approx \frac{NM \cdot b}{8} + \frac{NM}{G}(m_s + m_z),
\end{equation}
where $m_s$ and $m_z$ denote the number of bytes used to store the scale and zero-point for each group, respectively. This trade-off between adaptivity and metadata is one of the basic design choices in practical quantization systems.

\subsection{Quantization Targets in Transformer Models}
\label{sec:what_quantize}

The previous subsection described how quantization works at the level of individual values and matrices. We now turn to where quantization is applied in a Transformer. A Transformer-based foundation model consists of an embedding layer, a stack of $L$ repeated Transformer blocks, and an output head. In modern LLMs, most parameters are concentrated in the large linear projection matrices within the Transformer blocks, which is why quantization methods typically begin there. We use the LLaMA architecture~\citep{touvron2023llama,touvron2023llama2} as a representative example.

\subsubsection{Anatomy of a Transformer Block}
\label{subsubsec:anatomy-transformer}

A LLaMA-style Transformer is a decoder-only model composed of an embedding layer followed by a stack of $L$ pre-normalized Transformer blocks. Each block contains two main sub-layers: causal self-attention and a gated feed-forward network. Following standard practice, we omit biases and dropout for clarity, since the main targets of quantization are the large linear projection matrices.

For simplicity, let $X \in \mathbb{R}^{T \times d}$ denote the input to a Transformer block, where $T$ is the sequence length and $d$ is the hidden dimension. The block first applies RMS normalization before the attention sub-layer:
\begin{equation}
  X_{\mathrm{attn}} = \mathrm{RMSNorm}_{\mathrm{attn}}(X).
\end{equation}
The normalized input is then projected into queries, keys, and values:
\begin{equation}
  Q = \mathrm{RoPE}(X_{\mathrm{attn}} W_q), \qquad
  K = \mathrm{RoPE}(X_{\mathrm{attn}} W_k), \qquad
  V = X_{\mathrm{attn}} W_v,
  \label{eq:qkv}
\end{equation}
where $W_q, W_k, W_v \in \mathbb{R}^{d \times d}$ and $\mathrm{RoPE}(\cdot)$ denote rotary positional embeddings. After splitting $Q$, $K$, and $V$ into $H$ attention heads, we obtain
$Q^{(h)}, K^{(h)}, V^{(h)} \in \mathbb{R}^{T \times d_k}$ with $d_k = \nicefrac{d}{H}$. The output of head $h$ is
\begin{equation}
  A^{(h)} =
  \mathrm{softmax}\left(
    \frac{Q^{(h)} {K^{(h)}}^\top}{\sqrt{d_k}} + M_{\mathrm{causal}}
  \right)V^{(h)}
  \in \mathbb{R}^{T \times d_k},
  \label{eq:attention}
\end{equation}
where $M_{\mathrm{causal}} \in \mathbb{R}^{T \times T}$ is the causal mask with entries of $0$ for visible positions and $-\infty$ for masked future positions. The head outputs are then concatenated and projected back into the hidden dimension:
\begin{equation}
  X_{\mathrm{attn\_out}}
  = X + \mathrm{Concat}\bigl(A^{(1)}, \ldots, A^{(H)}\bigr) W_o,~~
  W_o \in \mathbb{R}^{d \times d}.
  \label{eq:attn_output}
\end{equation}
The block then applies a second RMS normalization before the feed-forward sub-layer:
\begin{equation}
  X_{\mathrm{ffn}} = \mathrm{RMSNorm}_{\mathrm{ffn}}(X_{\mathrm{attn\_out}}).
\end{equation}
In LLaMA, the feed-forward sub-layer employs a SwiGLU-style gated architecture with three weight matrices,
$W_{\mathrm{gate}}, W_{\mathrm{up}} \in \mathbb{R}^{d \times d_{\mathrm{ff}}}$ and
$W_{\mathrm{down}} \in \mathbb{R}^{d_{\mathrm{ff}} \times d}$:
\begin{equation}
  X_{\mathrm{out}}
  =
  X_{\mathrm{attn\_out}}
  +
  \left(
    \mathrm{SiLU}(X_{\mathrm{ffn}} W_{\mathrm{gate}})
    \odot
    (X_{\mathrm{ffn}} W_{\mathrm{up}})
  \right) W_{\mathrm{down}},
  \label{eq:ffn}
\end{equation}
where $\mathrm{SiLU}(x)=x \cdot \sigma(x)$ and $\odot$ denotes element-wise multiplication.

The main point for quantization is that each Transformer block contains several large dense projection matrices. In the LLaMA formulation above, there are seven such matrices per block: $W_q$, $W_k$, $W_v$, and $W_o$ in the attention sub-layer, and $W_{\mathrm{gate}}$, $W_{\mathrm{up}}$, and $W_{\mathrm{down}}$ in the feed-forward sub-layer. For a model with $L$ blocks, these contribute $7L$ large projection matrices, in addition to the token embedding matrix, the final normalization layer, and the output head. In practice, these large linear weights dominate both parameter count and memory footprint, making them the primary targets for weight quantization. Small normalization parameters and other auxiliary tensors are typically kept in higher precision.

For simplicity, the equations above use the standard multi-head attention form. Some LLaMA-family variants instead use grouped-query attention (GQA), in which the key and value projections have fewer heads than the query projection. This changes the shapes of $W_k$ and $W_v$, but it does not change the main quantization targets: the large attention and feed-forward projection matrices.

\subsubsection{Weight Quantization}

Among the possible quantization targets in a Transformer, weights are usually the natural starting point. In weight-only quantization, each linear projection matrix is stored in low precision while activations remain in floating-point format. During inference, specialized kernels either dequantize these weights on the fly or operate directly on low-precision representations, depending on the hardware and software stack. Because the overwhelming majority of Transformer parameters reside in large linear layers, quantizing weights alone often provides the largest immediate reduction in model memory footprint.

Consider a single linear layer with input activations $X \in \mathbb{R}^{T \times N}$ and a weight matrix $W \in \mathbb{R}^{N \times M}$, producing
\begin{equation}
  Y = XW \in \mathbb{R}^{T \times M}.
\end{equation}
Given a quantization set $\mathcal{Q}$ determined by the target bit-width, granularity, and calibration scheme, the goal is to find a quantized approximation $\widehat{W} \in \mathcal{Q}$. At the layer level, two formulations are especially common.

\paragraph{Direct Weight Quantization.}
The most straightforward objective is to keep the quantized weights numerically close to the full-precision weights:
\begin{equation}
  \widehat{W}
  =
  \argmin_{\widetilde{W} \in \mathcal{Q}}
  \|\widetilde{W} - W\|_F^2.
  \label{eq:direct_weight_quant}
\end{equation}
For a fixed quantization grid, this problem separates across individual entries, so each weight is mapped independently to the nearest representable value. This yields the standard round-to-nearest (RTN) quantizer, which is simple, computationally efficient, and often serves as a strong baseline. Its limitation is that it treats all perturbations equally: an error in a direction that barely changes the layer output is penalized just as heavily as an error in a direction to which the layer is highly sensitive.

\paragraph{Activation-Aware Quantization.}
A more faithful objective is to preserve the layer output under representative inputs. Instead of minimizing only weight distortion, activation-aware methods minimize the layer-wise output reconstruction error:
\begin{equation}
  \widehat{W}
  =
  \argmin_{\widetilde{W} \in \mathcal{Q}}
  \|X\widetilde{W} - XW\|_F^2
  =
  \argmin_{\widetilde{W} \in \mathcal{Q}}
  \|X(\widetilde{W} - W)\|_F^2.
  \label{eq:activation_aware_quant}
\end{equation}
Expanding this quadratic form reveals the role of the activation Gram matrix $X^\top X \in \mathbb{R}^{N \times N}$: it weights perturbations according to how strongly they affect the layer output. In the corresponding least-squares problem, $X^\top X$ captures the relevant second-order structure and is proportional, up to a constant factor, to the local Hessian. Unlike direct weight quantization, this objective acknowledges that some directions in weight space are far more sensitive than others.

This distinction is especially important in large language models, where two quantized weight matrices can be similarly close to $W$ in Frobenius norm yet induce very different output distortions because the layer is not equally sensitive in all directions. Activation-aware methods use calibration activations not only to determine quantization scales, but also to identify which weight perturbations matter most for preserving model behavior. GPTQ~\citep{frantar2022gptq}, for example, exploits this second-order structure to perform column-wise quantization with residual error compensation, substantially improving accuracy over naive RTN at low bit-widths.

In practice, direct weight quantization is attractive because it is simple, fast, and requires no activation statistics beyond those needed to define the quantization grid. Activation-aware quantization is more computationally demanding because it relies on calibration data and requires storing or approximating activation-dependent statistics. It nonetheless tends to deliver better accuracy, especially at 3--4 bits and below. As a result, many modern PTQ pipelines follow a natural progression: RTN provides a lightweight baseline, while activation-aware methods refine the quantized weights by explicitly accounting for output sensitivity.

\subsubsection{Activation Quantization}

Beyond weights, one can also quantize the intermediate tensors that flow through a Transformer block. From Eqs.~\eqref{eq:qkv}--\eqref{eq:ffn}, these include the normalized inputs, the query, key, and value tensors, the attention outputs, and the intermediate gated activations in the feed-forward network. Quantizing such tensors reduces activation memory and, when the hardware and kernels support it, enables low-precision matrix multiplication during inference. Unlike weights, however, activations are input-dependent and must be quantized dynamically as they are produced.

Activation quantization is generally more challenging than weight quantization for two main reasons. First, activation distributions can vary substantially across layers, channels, and input sequences, so a single static scale calibrated offline may be suboptimal at inference time. Second, Transformer activations often contain a small number of outlier channels whose magnitudes are much larger than those of typical activations~\citep{dettmers2022gpt3,czako2025activation_outlier}. Under uniform quantization, such outliers expand the quantization range and increase the step size $s$, forcing the majority of non-outlier values to be represented by relatively few discrete levels. The resulting loss of effective resolution can severely degrade model quality.

A representative solution is SmoothQuant~\citep{xiao2023smoothquant}, which applies per-channel rescaling to shift part of the quantization difficulty from activations to weights; see Section~\ref{sec:preprocess} for details. More broadly, activation quantization often requires careful calibration, dynamic scaling, or architectural transformations to control extreme dynamic ranges. When both activations and weights are quantized, the matrix multiplication $Y = XW$ can be executed with low-precision operands, typically using higher-precision accumulation followed by rescaling. On hardware with dedicated low-precision units, this can substantially improve inference efficiency~\citep{jacob2018quantization}.

\subsubsection{KV Cache Quantization}
\label{sec:act_kv_quant}

Autoregressive generation introduces another major memory term: the KV cache. To avoid recomputing attention over the entire prefix at every decoding step, Transformers store the key and value tensors produced at each layer for all previously generated tokens. For layer $l$ and KV head $h$, the cached tensors are precisely the key and value projections defined in Eq.~\eqref{eq:qkv}.

For a batch size of $B$, sequence length $T$, and a model with $L$ Transformer layers and $H_{\mathrm{kv}}$ key--value heads, the total KV cache contains $B L T H_{\mathrm{kv}} (d_k + d_v)$ elements, where $d_k$ and $d_v$ are the per-head dimensions of the key and value tensors. In most architectures, $d_k=d_v$, which reduces this expression to $2BLTH_{\mathrm{kv}}d_k$. The key point is that KV-cache memory grows linearly with sequence length $T$. For long contexts or large batch sizes, the cache can therefore become comparable to, or even larger than, the model weights themselves.

Quantizing cached keys and values to lower precision reduces this memory almost proportionally, enabling longer context windows, larger effective batch sizes, and higher serving throughput~\citep{ashkboos2024quarot}. In this paper, however, we focus on weight-only quantization; activation and KV-cache quantization are outside the scope of the present method.

\subsection{Post-Training Quantization and Quantization-Aware Training}
\label{sec:ptq_objective}

The previous subsection described what can be quantized. We now turn to the setting studied in this paper: post-training quantization. PTQ quantizes a pretrained model \emph{after} training has finished, without modifying the original optimization process~\citep{nagel2021white}. At a high level, PTQ seeks a quantized parameter set that makes the quantized model behave as similarly as possible to the original full-precision model on representative inputs.

Let $f_{\bm{\theta}}$ denote a full-precision model with parameters $\bm{\theta}$, and let $f_{\hat{\bm{\theta}}}$ denote its quantized counterpart with compressed parameters $\hat{\bm{\theta}} \in \mathcal{Q}$. A natural global PTQ objective is
\begin{equation}
  \min_{\hat{\bm{\theta}} \in \mathcal{Q}}
  \; \mathbb{E}_{x \sim \mathcal{P}}
  \left[
    d\!\left(f_{\bm{\theta}}(x), f_{\hat{\bm{\theta}}}(x)\right)
  \right],
  \label{eq:global_obj}
\end{equation}
where $d(\cdot,\cdot)$ measures the difference between the two model outputs and $\mathcal{P}$ denotes the underlying data distribution. This formulation simply states that the quantized model should stay close to the full-precision model on typical inputs.

In practice, however, PTQ does not have direct access to $\mathcal{P}$. Instead, it relies on a small calibration set $\mathcal{C} = \{\bm{x}_1, \ldots, \bm{x}_n\}$ sampled from $\mathcal{P}$, typically consisting of $n$ text sequences of length $T$. This calibration set is used to estimate the activation statistics needed to guide quantization. For each layer $l$, the calibration sequences are passed through the model to collect the input activation matrix $X_l \in \mathbb{R}^{nT \times N}$, where each row corresponds to a token-level activation vector. From $X_l$, one can compute layer-specific statistics such as the Gram matrix $H_l = X_l^\top X_l \in \mathbb{R}^{N \times N}$, which is closely related to the second-order information used by activation-aware PTQ methods. As prior work has shown, the quality of the calibration set has a direct effect on final quantization quality~\citep{pmlr-v139-hubara21a,williams-aletras-2024-impact}.

The appeal of PTQ lies in its practicality. It requires neither access to the original training data nor the full training pipeline, can be applied to essentially any pretrained checkpoint, and typically runs in minutes to hours rather than days. Its main challenge is that it must control quantization error without the benefit of end-to-end gradient-based adaptation. Consequently, the success of PTQ depends heavily on the design of effective local or global reconstruction objectives, robust calibration strategies, and error-compensation mechanisms. In this work, we focus on PTQ because it offers the strongest balance among applicability, computational cost, and compatibility with the rapidly growing ecosystem of open-weight models.

\paragraph{Relation to QAT.}
A complementary approach is \emph{quantization-aware training} (QAT)~\citep{jacob2018quantization,zhang2025qat_survey}, which incorporates simulated quantization directly into the training process. During each forward pass, weights and activations are quantized and then immediately dequantized, so the model learns in the presence of quantization noise. Because the rounding operator $\lfloor \cdot \rceil$ has zero gradient almost everywhere, QAT typically uses the straight-through estimator (STE)~\citep{bengio2013estimating}, which treats the gradient of rounding as the identity during backpropagation. This allows the model to adapt its parameters to compensate for quantization effects during training.

QAT can achieve excellent accuracy, especially at very low bit-widths, but it is substantially more expensive than PTQ. It requires access to the training pipeline and often incurs significant additional computation, in some cases hundreds of GPU-hours~\citep{liu2023llmqat}. Although recent work has demonstrated promising results for models up to 70B parameters, extending QAT reliably to 100B-scale models remains challenging. For this reason, PTQ remains the more practical and widely applicable setting for compressing modern foundation models.

\subsection{Practical PTQ Regimes: Layer-wise, Block-wise, and Global}
\label{sec:three_levels}

The PTQ objective in Eq.~\eqref{eq:global_obj} does not by itself determine how much of the model should be optimized jointly. In principle, the strongest formulation would optimize all quantization parameters over the entire model at once. In practice, the feasible scope of optimization is limited by available GPU memory during calibration and optimization. This leads to three common execution regimes, each corresponding to a different trade-off between optimization strength and memory cost. Figure~\ref{fig:workflow} provides an overview.

\textbf{Layer-wise PTQ}~\citep{frantar2022gptq,lin2024awq} quantizes one linear layer at a time and is the most memory-efficient regime. For each layer $l$, the method first obtains calibration activations $X_l$ by forwarding the calibration set through all preceding layers. It then solves a local reconstruction problem to determine the quantized weight $\widehat{W}_l$ for that layer, stores the result, and proceeds to the next layer. The peak GPU memory is bounded by a single weight matrix together with its associated Hessian-like statistic $H_l = X_l^\top X_l$, which makes this regime feasible even on a single consumer GPU. Layer-wise PTQ is also the most widely adopted setting in practice: many publicly available quantized models have been produced by tools such as AutoGPTQ~\citep{autogptq} and AWQ~\citep{awq_tool}.

The main limitation of layer-wise PTQ is that each layer is optimized in isolation. Once upstream layers have been quantized, downstream layers receive perturbed activations rather than the full-precision inputs assumed during local optimization. These discrepancies accumulate with depth, leading to \emph{error propagation}. This accumulation is one of the main reasons why layer-wise PTQ often lags behind more expensive optimization regimes in final model quality.

\textbf{Block-wise PTQ}~\citep{li2021brecq,ding2023cbq} jointly quantizes all linear layers inside one Transformer block at a time. A Transformer block is the model's basic repeating unit, typically consisting of self-attention, a feed-forward network, and normalization layers; modern LLMs often contain roughly 32 to 80 such blocks. Block-wise PTQ usually matches the output of the quantized block to that of its full-precision counterpart, allowing the optimizer to capture intra-block dependencies that layer-wise PTQ ignores, including interactions among the attention projections and the feed-forward layers. Blocks are processed sequentially, and the output of each quantized block becomes the input to the next.

\textbf{Global PTQ} is the most comprehensive regime. It loads the entire model onto GPUs and optimizes all quantization parameters jointly, often by minimizing an end-to-end divergence---for example, the KL divergence---between the full-precision teacher and the quantized student on the calibration set. Because the optimizer can coordinate adjustments across all layers simultaneously, this regime is potentially the most accurate. Its cost, however, is substantial: it requires enough GPU memory to hold both teacher and student models, together with optimizer states and cached activations.

The central idea behind \onecomp is that these three regimes need not be viewed as isolated alternatives. Instead, they can be organized into a coherent refinement pipeline. Layer-wise PTQ produces an initial quantized model, block-wise PTQ improves it, and global PTQ refines it further. Each stage builds on the output of the previous one, allowing users to stop at the point that best matches their hardware budget. A central contribution of \onecomp is to narrow the quality gap between layer-wise and global PTQ without requiring additional hardware; we return to this point in Section~\ref{sec:qep}.
\section{\onecomp Overview}
\label{sec:overview}

\onecomp is a resource-adaptive quantization framework that transforms a pretrained foundation model into a deployment-ready quantized model with minimal user intervention. 
It automatically selects the strongest quantization pipeline that fits
within the available GPU memory, from lightweight layer-wise methods
to full-model global optimization.
By primarily focusing on weight-only PTQ, it offers a compelling entry point for compression: quantizing weights offline significantly reduces memory footprint without altering the original training pipeline.
Its design is guided by three principles: an end-to-end workflow that automates major engineering steps, a pivot-based refinement architecture that supports progressively stronger post-quantization optimization, and a flexible, user-friendly interface for advanced use cases. This section offers a high-level overview of these design choices before detailing the technical aspects in subsequent sections.

\subsection{End-to-End Workflow}

Figure~\ref{fig:workflow} provides a high-level view of the \onecomp pipeline. Starting from a pretrained foundation model, \onecomp produces a quantized checkpoint through a sequence of automated stages. The pipeline first loads the target model and inspects its architecture, automatically identifying the model family, such as a dense large language model, a mixture-of-experts model, or a vision-language model, while detecting all quantizable linear layers and organizing them into the sequential processing order used by the quantization pipeline. This initial inspection enables the pipeline to operate across architectures without requiring users to manually specify the layer structure.

\subsubsection{Calibration Data Sampling}
\label{sec:autocalib}

All PTQ stages rely on a calibration dataset $\mathcal{C} = \{x_1, \ldots, x_n\}$ to estimate the activation statistics that guide quantization. These samples serve two related purposes. First, they provide the layer inputs $X_{l}$, from which \onecomp constructs the empirical input Gram matrix $H_l = X_l^\top X_l$, which serves as the per-row second-order term used by layer-wise PTQ methods. Second, they supply the reference outputs needed for block-wise and global reconstruction. Since quantization quality strongly depends on how well the calibration data reflect the target activation distribution~\citep{pmlr-v139-hubara21a}, \onecomp treats calibration as part of the default workflow rather than a manual preprocessing burden.

When users do not provide a custom calibration set, \onecomp automatically constructs one from a corpus. The default \texttt{concat\_chunk} strategy concatenates text samples, tokenizes the resulting text, and slices it into fixed-length chunks of $T$ tokens to produce $n$ calibration sequences. Alternative strategies, such as \texttt{drop\_head} and \texttt{drop\_rand}, reduce positional bias or increase sample diversity. This design keeps the default workflow simple while allowing advanced users to tailor calibration to a specific model or downstream setting.

Once the calibration data is prepared, \onecomp can also apply optional preprocessing transforms that make the model more amenable to low-bit compression. These include weight balancing through L1/L2 Sinkhorn normalization, incoherence transforms using Hadamard rotation, and channel equalization. Although these methods differ in form, they share the same goal: reducing outlier concentration to ensure subsequent quantization is more stable and less error-prone. Details are provided in Section~\ref{sec:preprocess}.

The core of the pipeline is the resource-adaptive quantization engine, which is organized as a sequence of refinement stages with increasing optimization scope. The pipeline always starts with \textbf{layer-wise PTQ}. In this regime, \onecomp loads one linear layer at a time onto the GPU, quantizes that layer, and then offloads the result to CPU memory. Peak memory is therefore bounded by a single weight matrix and a Hessian approximation computed from calibration activations, making this stage feasible even on a single consumer GPU. Importantly, this first stage already produces a self-contained quantized model that can be evaluated, saved, and deployed on its own.

When additional GPU memory is available, \onecomp uses this initial model as the starting point for \textbf{block-wise PTQ}. At this stage, the system loads one Transformer block at a time and jointly optimizes all quantization parameters within that block by distilling from the full-precision teacher; minimizing the discrepancy between the quantized and full-precision block outputs. With still larger resources, the pipeline can continue to \textbf{global PTQ}, which loads the full model and jointly optimizes quantization parameters across all layers, typically through KL-divergence-based distillation through minimizing the divergence between the quantized and full-precision model outputs. Finally, \onecomp can optionally perform parameter-efficient fine-tuning on top of a quantized backbone using QLoRA-compatible formats \citep{dettmers2023qlora} to further align the quantized model to a downstream task.

The key trade-off in Figure~\ref{fig:workflow} is between optimization scope and memory cost. Layer-wise PTQ operates on a single linear layer, block-wise PTQ on an entire Transformer block, and global PTQ on the full model. As the optimization unit grows, the method can correct more quantization error by accounting for interactions that smaller-scope methods must ignore, but it also requires substantially more GPU memory. \onecomp therefore adapts to the available hardware by advancing as far along this refinement chain as resources permit.

\subsection{Quantized Model as Pivot}

A central architectural decision in \onecomp is to treat the first quantized model produced by layer-wise PTQ as the pivot for all subsequent improvements. Rather than viewing layer-wise, block-wise, and global PTQ as unrelated modes, \onecomp organizes them into a progressive refinement pipeline. Layer-wise PTQ always runs first because it has the lowest memory requirement and already yields a usable quantized model. Every later stage---including block-wise refinement, global refinement, and optional fine-tuning---is implemented as a post-processing step that takes a quantized model as input and returns an improved one.

This pivot-based design, illustrated in Code~\ref{code:refiner}, has three practical advantages. First, it keeps the entry point simple: users who only need a reasonable quantized model can stop after the initial PTQ stage. Second, it facilitates smooth and incremental quality improvements: users can enable stronger refinement stages one at a time and directly measure the gains obtained from each additional computation budget. Third, it makes the framework naturally extensible. Any post-quantization technique that maps one quantized model to a better one can be integrated as a new refiner without changing the rest of the pipeline. In this sense, \onecomp serves not only as a quantization method but also as a composable framework in which new post-quantization techniques can be integrated as additional refiners without modifying
the existing pipeline..

\begin{codefloat}[t]
    \begin{codebox}
\begin{lstlisting}[style=onecomp-python]
# Step 1: Layer-wise PTQ (always runs, single GPU)
quantized_model = layerwise_quantizer.run(fp_model)

# Step 2+: Optional refinement pipeline
refinement_pipeline = [
    BlockwiseRefiner(...),    # block-level MSE distillation
    GlobalRefiner(...),       # model-level KL refinement
    FinetuningRefiner(...),   # QLoRA / adapter training
]

for refiner in refinement_pipeline:
    quantized_model = refiner.run(quantized_model)
\end{lstlisting}
    \end{codebox}
    \caption{Composable refinement pipeline.
    Each refiner accepts a quantized model and returns an improved one.}
    \label{code:refiner}
\end{codefloat}

\subsection{User Interface}

The architectural choices above are revealed through a deliberately minimal user interface. As shown in Code~\ref{code:api}, the user specifies the target model and launches the pipeline with a single high-level invocation. By default, \onecomp handles model inspection, calibration construction, quantization, evaluation, and checkpoint export automatically. This enables non-expert users to obtain a strong, deployment-ready baseline without needing to consider layer grouping, memory scheduling, or calibration mechanics.

At the same time, the interface is not restrictive. For researchers and advanced practitioners, each major decision point---including preprocessing, calibration, optimization regime, and refinement strategy---can be overridden through optional arguments. The result is a two-level interface: a simple default path for users who want the system to make sensible choices automatically, and an expert mode for users who wish to control the individual components of the pipeline. This separation is intentional. It keeps the common use case lightweight while ensuring that \onecomp remains a useful experimental platform for developing and evaluating new quantization methods.

\begin{codefloat}[tb]
\begin{codebox}
\begin{lstlisting}[style=onecomp-python]
uv run onecomp <model_path>
\end{lstlisting}
\end{codebox}
\caption{Minimal usage of \onecomp.
A single high-level invocation quantizes the target model, evaluates the result, and saves the resulting checkpoint.}
\label{code:api}
\end{codefloat}
\section{Quantization Engines}
\label{sec:quantization}

This section details the method underlying each stage of the \onecomp pipeline, which was introduced at a high-level in Section~\ref{sec:overview}.
We first describe two preprocessing stages; automatic mixed-precision planning (Section~\ref{sec:autobit}) and equivalent transformation (Section~\ref{sec:preprocess}).
Next, we describe three quantization regimes that increase the optimization target; layer-wise PTQ (Section~\ref{sec:layerwise}), block-wise PTQ (Section~\ref{sec:blockwise}), and global PTQ (Section~\ref{sec:global}).
Section~\ref{sec:finetuning} addresses optional quantization fine-tuning as a final refinement step.

\subsection{\autobit: Automatic Mixed-Precision Planning}
\label{sec:autobit}

Transformer layers differ markedly in their sensitivity to quantization. Early layers and attention output projections are often particularly fragile, while MLP layers in the middle of the network typically accommodate much more aggressive compression. As a result, assigning a uniform bit-width across all layers, as many existing tools do, misallocates precision: it overprovisions robust layers while underprovisioning the most sensitive ones. To address this imbalance, \autobit explicitly leverages layerwise heterogeneity by solving a constrained optimization problem that selects the bit-width for each layer or module under a global memory budget before quantization is performed.

\autobit first determines the target average bit-width, which serves as the effective global budget based on the user-specified VRAM constraint. If this value falls below a predefined threshold, such as 2 bits, the current release does not support mixed-bit allocation in such low-bit regimes; however, support for this scenario, along with DBF described in Section~\ref{sec:dbf}, is planned for a future release. 
Otherwise, \autobit solves an integer linear optimization problem to determine the bit-width allocation for each module.
Let $\mathcal{C}_l$ denote the set of candidate configurations for layer $l$, where each candidate $c \in \mathcal{C}_l$ specifies a bit-width, numeric format, and group size.
Each candidate has a memory cost $\mathrm{cost}(l, c)$ (the packed weight size plus per-group scale and zero-point overhead) and a predicted quantization error $\mathrm{err}(l, c)$.
Given a total budget $\mac{C^*}$, \autobit solves:
\begin{equation}
  \min_{\{c_l \in \mathcal{C}_l\}}
  \sum_{l=1}^{L} \mathrm{err}(l, c_l)
  \quad \text{s.t.} \quad
  \sum_{l=1}^{L} \mathrm{cost}(l, c_l) \le \mac{C^*},
  \label{eq:autobit}
\end{equation}
The \emph{module-level solver} formulates Eq.~\eqref{eq:autobit} as an integer linear programming problem and can be solved by using the SCIP\footnote{\url{https://www.scipopt.org}} solver in OR-Tools.
This enables finer-grained allocation, allowing different modules within the same layer to receive varying bit-widths.

To accurately reflect the impact of quantization on model performance, estimating the error metric $\mathrm{err}(l, c_l)$ is critical. Let $\Delta W^{(l, c_l)} \coloneqq \widehat{W}^{(l, c_l)} - W^{(l)}$ denote the weight perturbation obtained by running a fast round-to-nearest (RTN) proxy under configuration $c_l$.
It can be parallelly calculated per module and configuration. 
A naive approach measures $\mathrm{err}(l, c_l) = \|\Delta W^{(l, c_l)}\|_F^2$, but it ignores activation distributions.
We employ an activation-aware error based on the second-order Taylor expansion of the per-layer loss, expressed using the Kronecker-factored Hessian:
\begin{equation}
    \mathrm{err}(l, c_l) \approx \frac{1}{2} {\rm tr}\left( B^{(l)} \Delta W^{(l, c)} A^{(l)} {\Delta W^{(l, c)}}^{\top} \right)
\end{equation}
where $A^{(l)}$ and $B^{(l)}$ are the input Gram matrix and output curvature for layer $l$, respectively.
$\Delta W^{(l, c)}$ can be obtained by running the fast round-to-nearest (RTN) proxy.
Note that when we set $A^{(l)}$ and $B^{(l)}$ as the identity matrices, the increase of loss is equivalent to the naive Frobenius norm.
In practice, we consider only the diagonal elements of $A^{(l)}$ and $B^{(l)}$ for saving computational budget.

\subsection{Preprocessing via Equivalent Transformations}
\label{sec:preprocess}

Transformer models often exhibit heavy-tailed distributions in both weights and activations, with a small subset of channels containing values that are orders of magnitude larger than the rest~\citep{dettmers2022gpt3,czako2025activation_outlier}. These outliers increase the quantization step size $s$, effectively reducing the resolution available for typical values and degrading the overall model accuracy. To mitigate this issue, preprocessing methods employ mathematically equivalent transformations to redistribute outlier energy prior to quantization, without altering the underlying function represented by the model.

\subsubsection{Channel-wise Scaling}

SmoothQuant~\citep{xiao2023smoothquant} introduces a per-channel scaling vector $\bm{s} \in \mathbb{R}^N$ applied jointly to activations and weights:
\begin{equation}
    Y = XW = \underbrace{(X\mathrm{diag}(\bm{s}))}_{\widetilde{X}} \cdot \underbrace{(\mathrm{diag}(\bm{s})^{-1} W)}_{\widetilde{W}},
  \label{eq:smoothquant}
\end{equation}
This transformation preserves exact equivalence since $\mathrm{diag}(\bm{s})^{-1}\mathrm{diag}(\bm{s}) = I_N$. The scaling factors are defined as $s_j = \nicefrac{\max(|X_j|)^\alpha}{\max(|W_j|)^{1-\alpha}}$, 
where $\alpha \in [0,1]$ controls the trade-off in quantization difficulty between activations and weights. By redistributing magnitude across the two, SmoothQuant reduces extreme values in either domain.
OmniQuant~\citep{shao2023omniquant} generalizes this concept by introducing learnable equivalent transformations, including optimized scaling and shifting parameters that minimize block-wise quantization error. This formulation allows for greater flexibility in adapting to layer-specific statistics.

\subsubsection{Rotation-based Transformations}

While channel-wise scaling effectively mitigates persistent channel outliers, it is less effective against token-specific activation spikes that occur in only a few channels, often referred to as massive outliers~\citep{czako2025activation_outlier}. Rotation-based methods address this limitation by applying an orthogonal transformation $R \in \mathbb{R}^{N \times N}$:
\begin{equation}
  Y = XW = \underbrace{(XR)}_{\widehat{X}} \cdot \underbrace{(R^\top W)}_{\widehat{W}}, 
  \label{eq:rotation}
\end{equation}
where the orthogonality constraint $RR^{\top}=I_{N}$ ensures exact functional equivalence. Intuitively, orthogonal transformations preserve vector norms while redistributing energy across dimensions. This reduces the concentration of large values in individual coordinates and enhances quantization.

QuIP~\citep{chee2024quip} formalizes this intuition through the concept of incoherence. A matrix $W \in \mathbb{R}^{N \times M}$ is $\mu$-incoherent if
\begin{equation}
  \max_{i,j} |W_{ij}| \le \mu \|W\|_F / \sqrt{NM}.
\end{equation}

Unlike fixed or random rotations, SpinQuant optimizes rotation matrices directly on the orthogonal group. QuIP\#~\citep{tseng2024quip} enhances this approach by replacing dense random orthogonal matrices with efficient randomized Hadamard transforms and introducing lattice-based codebooks. QuaRot~\citep{ashkboos2024quarot} unifies rotation-based preprocessing for hidden states and attention-related activations, including KV caches, to enable end-to-end improvements in quantization.

\onecomp incorporates the SpinQuant~\citep{liu2024spinquant} framework, which learns task-adaptive orthogonal transformations.
\begin{equation}
  \mathrm{O}(N) = \{R \in \mathbb{R}^{N \times N} : R^{\top} R = I_N\}
\end{equation}
using CayleySGD~\citep{li2020efficient}, which updates orthogonal matrices while preserving orthogonality. Compared to random or fixed structured rotations, learned rotations adapt more effectively to a model's statistics and significantly enhance quantization quality. After optimization, mergeable rotations can be folded offline into surrounding linear weights, while a small number of non-mergeable transforms are retained as efficient online Hadamard operations. 
\onecomp provides two categories of rotation-based preprocessing.
As a standalone model-wide preprocessing step, \onecomp incorporates SpinQuant~\citep{liu2024spinquant}, which learns task-adaptive orthogonal transformations on $\mathrm{O}(N) = \{R \in \mathbb{R}^{N \times N} : R^{\top} R = I_N\}$ using CayleySGD~\citep{li2020efficient}.
After optimization, mergeable rotations are folded offline into surrounding linear weights, while non-mergeable transforms are kept as efficient online Hadamard operations.
In addition, \onecomp supports QuIP-style per-layer random rotations~\citep{chee2024quip} within its layer-wise PTQ path (Section~\ref{sec:layerwise}), where the rotation and quantization are performed jointly in the same precision.

\onecomp also supports L1/L2 Sinkhorn weight balancing, which iteratively rescales rows and columns to achieve more uniform magnitudes across both dimensions. This reduces the dynamic range the quantizer must represent and enhances quantization accuracy while keeping the layer output unchanged.

\subsection{Layer-wise PTQ}
\label{sec:layerwise}

\begin{figure}[t]
  \centering
  \includegraphics[width=\linewidth]{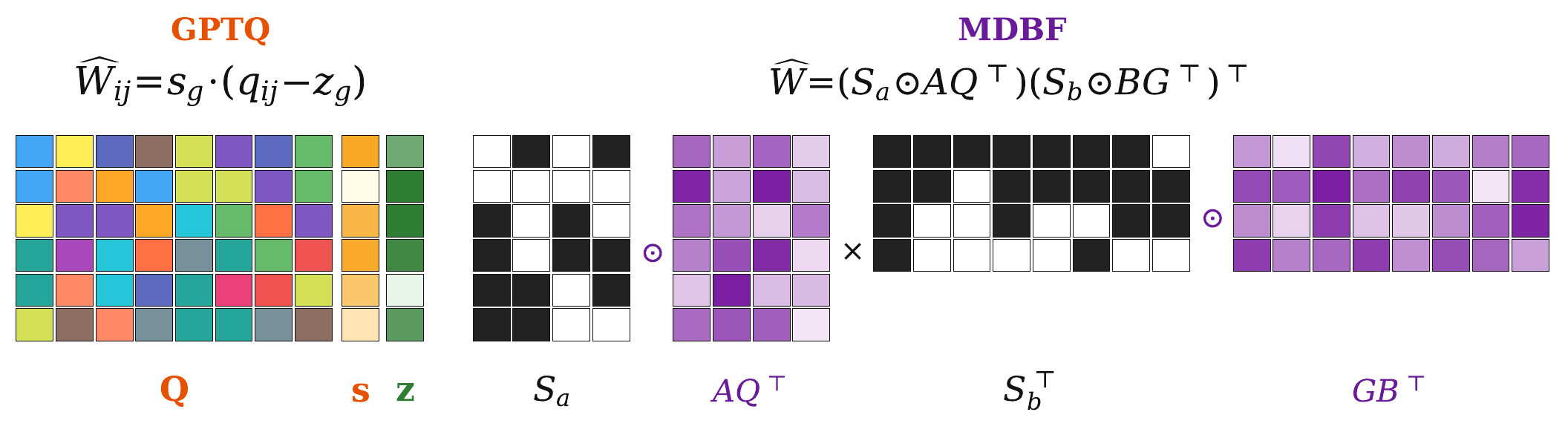}
  \caption{Two quantization formats supported by \onecomp.
  Left: GPTQ stores per-group integer weights $\bm{Q}$, scales $\bm{s}$, and zero-points $\bm{z}$, and reconstructs weights via $\hat{W}_{ij} = s_g(q_{ij} - z_g)$.
  Right: MDBF represents the weight matrix using shared binary sign bases $S_a, S_b$ together with low-rank real-valued envelopes $AP^\top$ and $BG^\top$, yielding $\hat{W} = (S_a \odot AP^\top)(S_b \odot BG^\top)^\top$.}
  \label{fig:formats}
\end{figure}

Layer-wise PTQ quantizes a model one linear layer at a time.
We begin by introducing the notation used throughout this section.
Let $W_l \in \mathbb{R}^{N \times M}$ denote the full-precision weight matrix of layer $l$, with $N$ and $M$ representing the input and output dimensions, respectively.
Let $X_l \in \mathbb{R}^{T \times N}$ be the matrix of $T$ calibration activations collected from the full-precision model at layer $l$, with the corresponding full-precision output being $Y_l = X_l W_l \in \mathbb{R}^{T \times M}$.
We denote $\widehat{X}_l \in \mathbb{R}^{T \times N}$ as the activations produced at the same layer by the incrementally quantized model after quantizing all preceding layers.

An activation-aware layer-wise PTQ objective aims to find a quantized weight matrix $\widehat{W}_l$ that minimizes the output reconstruction error based on the calibration dataset:
\begin{equation}
  \widehat{W}_l = \argmin_{\widehat{W} \in \mathcal{Q}} \left\| \widehat{X}_l(\widehat{W} - W_l) \right\|_F^2,
  \label{eq:layerwise}
\end{equation}
where $\mathcal{Q} \subset \mathbb{R}^{N \times M}$ denotes the set of quantized weight matrices allowed by the selected quantization format.
The Gram matrix, 
$\widehat{H}_l = \widehat{X}_l^{\top} \widehat{X}_l \in \mathbb{R}^{N \times N}$,
is a positive semidefinite matrix that captures the relative importance of different input directions.
GPTQ~\citep{frantar2022gptq} approximately solves Eq.~\eqref{eq:layerwise} column by column, using second-order information from $\widehat{H}_l$; at each step, it quantizes one column and propagates the resulting residual to the remaining unquantized columns.
By default, columns are processed in their natural order; an optional activation ordering (\texttt{actorder}) permutes columns based on the decreasing magnitude of the Hessian diagonal, ensuring that the most sensitive columns are quantized first when the capacity for residual compensation is greatest.
Independently, per-group quantization scales can be set either by simple min-max calibration or by MSE-based grid tuning (\texttt{mse}), which searches over candidate scales to minimize the mean squared reconstruction error per-group.
AWQ~\citep{lin2024awq} emphasizes significant channels identified from activation statistics and rescales them before rounding.
A key advantage of layer-wise PTQ is its efficiency; it processes one layer at a time, resulting in a small memory footprint and enabling the rapid quantization of large-scale models using only a single GPU.
Its lightweight nature also makes layer-wise PTQ a practical initialization strategy for more expensive end-to-end PTQ pipelines.

The central challenge in layer-wise PTQ, however, is ``error propagation''.
Since upstream layers have already been quantized, layer $l$ receives perturbed inputs $\widehat{X}_l \neq X_l$, causing the errors introduced early in the network to accumulate as they pass through subsequent layers.
\onecomp is designed to preserve the computational efficiency and single-GPU practicality of layer-wise PTQ while incorporating mechanisms to explicitly mitigate quantization error propagation.
As a result, it retains the lightweight advantages of layer-wise PTQ while enhancing robustness against accumulated upstream quantization errors.

More broadly, improving layer-wise PTQ involves two complementary aspects.
The first is the optimization procedure: how the quantization objective is addressed, particularly in the presence of accumulated upstream errors.
The second is the quantization representation: what format the quantized weights take and how expressive that format is at a given bit budget.
\onecomp addresses both aspects; its optimization techniques mitigate error propagation and cross-layer inconsistencies, while its representation choices range from improved uniform formats to structured binary factorizations for extreme low-bit regimes.

\subsubsection{Optimization \label{sec:lptq_opt}}
\onecomp incorporates two optimization techniques that go beyond standard layer-wise quantization.
\qep ~corrects the quantization target of each layer to compensate for errors introduced by preceding quantized layers, while \lpcd ~extends the optimization scope from individual layers to coupled submodules within each transformer block.
Both techniques are compatible with any base layer-wise quantizer and add only modest computational overhead.

\paragraph{\qep: Quantization Error Propagation}\label{sec:qep}

\qep~\citep{arai2025quantization} improves layer-wise PTQ by modifying the quantization target before applying the base quantizer.
Its central observation is that, once upstream layers are quantized, the current layer no longer operates on the original activations.
Thus, quantizing the current layer to best match the original full-precision weights is generally suboptimal.
A better approach is to adjust the target weights to compensate for errors introduced by preceding layers.

To this end, instead of minimizing the standard activation-aware objective in Eq.~\eqref{eq:layerwise}, \qep seeks a quantized weight matrix that reconstructs the full-precision output $X_l W_l$ from the perturbed input $\widehat{X}_l$:
\begin{equation}
  \widehat{W}^{\mathrm{QEP}} = \argmin_{\widehat{W}} \bigl\| \widehat{X}_l \widehat{W} - X_l W_l \bigr\|_F^2.
  \label{eq:qep_obj}
\end{equation}

Define the activation error as $\bm{\delta}_l = X_l - \widehat{X}_l \in \mathbb{R}^{T \times N}$, and the corresponding error-propagation matrix as $\bm{C}_l = \widehat{X}_l^\top \bm{\delta}_l \in \mathbb{R}^{N \times N}$.
As shown by \citet{arai2025quantization}, the unconstrained minimizer of Eq.~\eqref{eq:qep_obj} is given by
$\overline{W} = (I_N + \widehat{H}_l^{-1}\bm{C}_l)W_l$.
Therefore, the QEP objective can be interpreted as the standard layer-wise PTQ problem in Eq.~\eqref{eq:layerwise}, but with a corrected quantization target.

In practice, \qep introduces a propagation-strength parameter $\alpha_l \in [0,1]$ to interpolate between the original weight matrix and the fully corrected target.
To improve numerical stability, it also applies Tikhonov regularization:
\begin{equation}
  W_l^\star(\alpha_l) = \bigl(I_N + \alpha_l (\widehat{H}_l + \lambda I_N)^{-1} \bm{C}_l \bigr) W_l,
  \label{eq:qep}
\end{equation}
where $\lambda = \eta \cdot \mathrm{mean}(\mathrm{diag}(\widehat{H}_l))$ and $\eta > 0$ is a small constant.
The regularized inverse is efficiently computed using Cholesky factorization of $\widehat{H}_l + \lambda I_N$.

The corrected target $W_l^\star$ can then be passed to any base quantizer, making \qep a straightforward plug-and-play enhancement for layer-wise PTQ.
During the calibration forward pass, \onecomp maintains two parallel model instances, accumulating $\widehat{H}_l$ and $\bm{C}_l$ online and avoiding the need to store full activation matrices.
In addition to this layer-wise formulation, another extension, LoaQ \citep{lin2025loaq}, has been proposed to enable quantization at the submodule level.
\onecomp supports both \qep and its submodule-level extension LoaQ, providing a unified framework for error-propagation-aware quantization across various granularities.

\paragraph{\lpcd: Submodule-Aware Coordinate Descent}
\label{sec:lpcd}

While \qep mitigates error propagation at the individual linear layer level, transformer computation is fundamentally structured around submodules that couple multiple layers into a single functional unit.
The query and key projections determine the attention score matrix, while the value and output projections dictate the attention output; the up and down projections collaboratively establish the MLP output.
As a result, quantizing these coupled layers independently can introduce cross-layer inconsistencies that degrade the overall performance of the submodule.

\lpcd~\citep{ichikawa2025lpcd} addresses this issue by extending PTQ from individual layers to arbitrary submodules while maintaining the efficiency of layer-wise quantization pipelines.
Its key insight is that independently quantizing interacting layers can be suboptimal, because the optimal quantized value for one layer generally depends on the quantized values of its coupled modules.
To account for this dependency, LPCD jointly optimizes groups of interacting layers through an iterative relax-then-project procedure.
At each step, it first computes the optimal continuous-valued solution for one layer while keeping the other layers in the submodule fixed at their current quantized values, and then projects that solution back onto the quantized domain using a standard layer-wise quantizer.
Thus, LPCD inherits the strengths and flexibility of existing layer-wise PTQ methods while allowing for coordination at the submodule-level.

Formally, consider a submodule with $R$ weight matrices $\{M_1, \ldots, M_R\}$ with $M_r \in \mathbb{R}^{N_r \times K_r}$, and let $\widehat{M}_r \in \mathcal{Q}$ denote their quantized counterparts.
Let $\mathcal{J}(\widehat{M}_1, \ldots, \widehat{M}_R)$ measure the discrepancy between the outputs of the full-precision and quantized submodules.
The resulting discrete optimization problem is:
\begin{equation}
  \min_{\widehat{M}_1 \in \mathcal{Q}, \ldots, \widehat{M}_R \in \mathcal{Q}} \mathcal{J}(\widehat{M}_1, \ldots, \widehat{M}_R).
  \label{eq:lpcd_global}
\end{equation}
Since the feasible set is discrete and highly non-convex, solving this problem is intractable.
\lpcd approximates the solution using cyclic block coordinate descent.
In outer iteration $t$, the blocks are updated sequentially for $r = 1, \ldots, R$. Each block update consists of two stages.

In the relaxation step, the quantization constraint on $M_r$ is temporarily lifted, allowing the continuous solution to be computed while all other blocks remain fixed at their most recently updated quantized values:
\begin{equation}
  \overline{M}_r^{(t)} = \argmin_{U \in \mathbb{R}^{N_r \times K_r}} \mathcal{J}(\widehat{M}_1^{(t)}, \ldots, \widehat{M}_{r-1}^{(t)}, U, \widehat{M}_{r+1}^{(t-1)}, \ldots, \widehat{M}_R^{(t-1)}).
  \label{eq:lpcd_relax}
\end{equation}
For the submodules considered in this work, $\mathcal{J}$ is quadratic in $U$, making this relaxation admit a closed-form least-squares solution.
In practice, however, the design matrices involved are often too large for direct solutions; therefore, we approximate the optimum by optimizing the original loss using Adam.
In the subsequent projection step, the continuous solution $\overline{M}_r^{(t)}$ is mapped back to the quantized set using a standard layer-wise PTQ quantizer $\Pi_\mathcal{Q}$:
\begin{equation}
  \widehat{M}_r^{(t)} \leftarrow \Pi_\mathcal{Q}(\overline{M}_r^{(t)}).
  \label{eq:lpcd_project}
\end{equation}
The projector $\Pi_\mathcal{Q}$ can be instantiated with any existing layer-wise quantizer, such as GPTQ, AWQ, RTN, or DBF, making LPCD immediately compatible with broader layer-wise PTQ methods.
An important theoretical result of \citet{ichikawa2025lpcd} is that QEP arises as a special case of a single LPCD iteration applied to the two-block objective
$\mathcal{J}(\widehat{W}, \widehat{X}) = \|\widehat{X}\widehat{W} - XW\|_F^2$,
with the activation block held fixed.
LPCD generalizes this perspective by allowing multiple coordinate-descent iterations and arbitrary submodule decompositions; we refer the reader to \citet{ichikawa2025lpcd} for complete theoretical details.

\onecomp instantiates LPCD with three submodule groups in each transformer layer, each corresponding to a distinct functional interaction.
The \emph{QK module} $\{W_q, W_k\}$ collaboratively optimizes the query and key projections to minimize the attention-score reconstruction error.
The \emph{VO module} $\{W_v, W_o\}$ jointly optimizes the value and output projections to minimize the attention-output error while considering the impact of the residual stream.
The \emph{gate-up-down module} $\{W_{\mathrm{gate}}, W_{\mathrm{up}}, W_{\mathrm{down}}\}$ jointly optimizes the MLP projections to minimize the MLP-output reconstruction error.
For each submodule, LPCD iterates over the weight matrices, alternating between relaxation and projection steps.
In practice, convergence is typically reached within 2--4 outer iterations, and the additional cost remains modest relative to the runtime of the underlying layer-wise quantization procedure.

\subsubsection{Representation \label{sec:lptq_representation}}
Beyond the optimization procedure, a key design choice in layer-wise PTQ is the quantized representation to which the weights are mapped.
\onecomp supports two families of quantization formats.
\jointq refines the conventional per-group uniform format used by GPTQ through joint optimization of scales and integer weights, targeting the 3--4 bit regime.
For more aggressive compression at 1--2 bits, \onecomp provides DBF and its generalization MDBF, which replace the uniform grid with structured binary sign factorizations equipped with learnable magnitude envelopes.

\paragraph{\jointq: Joint Scale-and-Integer Optimization}
\label{sec:jointq}

We propose \jointq adopting the same quantization format as GPTQ~\citep{frantar2022gptq}, namely layer-wise uniform scalar quantization with group-wise scales and zero-points, in either the symmetric or asymmetric setting; it solves the following optimization problem over group scales $s_t$, group zero-points $z_t$, and integer weights $q_{ij}$:
\begin{equation}
  \begin{aligned}
    \min \quad & \bigl\|X W - X\widehat{W}\bigr\|_F^2 \\
    \text{s.t.} \quad & \widehat{W}_{ij} = s_{g(i,j)}\bigl(q_{ij} - z_{g(i,j)}\bigr) \quad \forall i,j,\\
    &q_{ij} \in Q, \quad \forall i,j,\\
    &s_{t} \in \mathbb{R}, \quad z_{t} \in Q, \quad \forall t.
  \end{aligned}
  \label{eq:jointq}
\end{equation}
Here $g(i,j)$ maps a weight element $(i,j)$ to its group index, for example, a contiguous block of size $d$ within each row. The integer codebook $Q\subset\mathbb{Z}$ is determined by the bit width $b$ and whether the quantization scheme is symmetric or asymmetric; for instance, for 4-bit symmetric quantization, $Q=[-8, 7]\cap\mathbb{Z}$. In the symmetric case, all zero-points are fixed to zero, i.e., $z_t=0$ for all $t$.

\jointq solve this problem using a local-search-based algorithm. For each row $i$, the algorithm repeatedly updates the integer weights $q_{ij}$ along with the corresponding group scale $s_{g(i,j)}$ to decrease the objective in Eq.~\eqref{eq:jointq}. Concretely, each local move proposes a small change to one integer weight $q_{ij}$ within the codebook $Q$; conditioned on this discrete change, the scale of the affected group is updated in the same step to best fit the current integer assignments. The proposed move is accepted if it improves the objective, and the procedure terminates when no improving local move can be found. A key design choice of \jointq is to jointly optimize the scales and integer weights rather than optimizing them in separate alternating subproblems.

\jointq can optionally incorporate a proximity regularizer to discourage excessive deviation from the original weights:
\begin{equation}
  \min \ \bigl\|X W - X\widehat{W}\bigr\|_F^2 + n\lambda \bigl\|W-\widehat{W}\bigr\|_F^2,
  \label{eq:jointq_reg}
\end{equation}
where $n$ is the number of rows of $X$ and $\lambda>0$ is a hyperparameter. Expanding the objective shows that Eq.~\eqref{eq:jointq_reg} is equivalent to Eq.~\eqref{eq:jointq} with an augmented input matrix, hence the same local-search procedure can be applied without modification.
Finally, since \jointq uses the same quantization format as GPTQ, a GPTQ-quantized model can be directly used as an initial solution. The output of \jointq is a standard GPTQ-format checkpoint compatible with existing inference kernels.

\paragraph{DBF and MDBF: Extreme Low-Bit Formats}
\label{sec:dbf}

\onecomp is a unified post-training quantization package that supports extremely low-bit compression across a wide range of bit widths.
To this end, it provides efficient quantization pipelines for both conventional uniform formats and more expressive structured formats tailored to the extreme low-bit regime.
Figure~\ref{fig:formats} compares the two quantization formats supported by the package: the per-group uniform GPTQ format for 3--4 bit quantization and the structured binary-factor MDBF format for the more challenging 1--2 bit regime.
At such extreme bit widths, uniform quantization grids often lack the expressive power needed to accurately represent the underlying weight distribution.
DBF~\citep{boza2026addition} addresses this limitation by factorizing each weight matrix into two binary sign matrices interleaved with diagonal scaling matrices:
\begin{equation}
  \widehat{W}_{l} = D_{\bm{a}^{(l)}} S_{a}^{(l)} D_{\bm{m}^{(l)}} (S_b^{(l)})^{\top} D_{\bm{b}^{(l)}},
  \label{eq:dbf}
\end{equation}
where $S_a^{(l)} \in \{\pm1\}^{N \times R}$ and $S_b^{(l)} \in \{\pm1\}^{M \times R}$ are binary sign matrices, and $D_{\bm{a}^{(l)}}, D_{\bm{m}^{(l)}}, D_{\bm{b}^{(l)}}$ are diagonal scaling matrices parameterized by the vectors $\bm{a}^{(l)} \in \mathbb{R}^N$, $\bm{m}^{(l)} \in \mathbb{R}^R$, and $\bm{b}^{(l)} \in \mathbb{R}^M$.
This representation can be viewed as a structured rank-$R$ factorization $\widehat{W} = \widehat{U}^{(l)}(\widehat{V}^{(l)})^\top$, where $\widehat{U}^{(l)} = D_{\bm{a}^{(l)}} S_a^{(l)} D_{(\bm{m}^{(l)})^{1/2}}$ and $\widehat{V}^{(l)} = D_{\bm{b}^{(l)}} S_b^{(l)} D_{(\bm{m}^{(l)})^{1/2}}$.
At inference time, DBF replaces a single high-precision GEMMs with two binary GEMMs interleaved with inexpensive diagonal rescalings, resulting in a hardware-efficient execution path.

Despite this advantage, DBF has a fundamental structural limitation.
After factoring out the binary sign bases, the demodulated envelopes
$E_{S_a^{(l)}}(\widehat{U}^{(l)}) = S_a^{(l)} \odot \widehat{U}^{(l)} = \bm{a}^{(l)}((\bm{m}^{(l)})^{1/2})^\top$
and
$E_{S_b^{(l)}}(\widehat{V}^{(l)}) = S_b^{(l)} \odot \widehat{V}^{(l)} = \bm{b}^{(l)}((\bm{m}^{(l)})^{1/2})^\top$
are each constrained to rank one.
Consequently, all $R$ columns of a factor share the same magnitude profile, up to a scalar multiplier.
Increasing $R$ primarily enhances the diversity of sign patterns while providing limited additional capacity for modeling magnitudes, which can lead to accuracy saturation.
MDBF~\citep{ichikawa2025mdbf} addresses the single-envelope bottleneck by replacing the rank-one magnitude envelope with a rank-$l$ envelope while preserving both the shared binary sign bases and the deployment-friendly binary inference primitive:
\begin{equation}
  \widehat{W}_l = \widehat{U}^{(l)} (\widehat{V}^{(l)})^\top,~~~
  \widehat{U}^{(l)} = S_a^{(l)} \odot (A^{(l)}(Q^{(l)})^\top),~~
  \widehat{V}^{(l)} = S^{(l)}_b \odot (B^{(l)}(G^{(l)})^\top),
  \label{eq:mdbf}
\end{equation}
where $A^{(l)} \in \mathbb{R}^{N \times l}$, $Q^{(l)} \in \mathbb{R}^{R \times l}$, $B^{(l)} \in \mathbb{R}^{M \times l}$, and $G^{(l)} \in \mathbb{R}^{R \times l}$ are real-valued factor matrices.
Under this formulation, the demodulated envelopes satisfy
$\mathrm{rank}(E_{S^{(l)}_a}(\widehat{U}^{(l)}_l)) = \mathrm{rank}(A^{(l)}(Q^{(l)})^\top) \le l$
and
$\mathrm{rank}(E_{S_b^{(l)}}(\widehat{V}^{(l)}_l)) = \mathrm{rank}(B^{(l)}(G^{(l)})^\top) \le l$,
allowing the model to represent multiple magnitude modes.
This result highlights a key distinction between MDBF and DBF: MDBF focuses its limited representational capacity on modeling magnitudes rather than merely increasing the diversity of sign patterns.
This allocation is especially advantageous in the extreme low-bit regime, where magnitude expressiveness becomes the critical bottleneck.
\onecomp adopts a two-stage optimization pipeline for MDBF.
It first computes a closed-form initialization using Multi-Envelope SVID, applying truncated SVD to the entrywise absolute values of each factor, followed by alternating ADMM-based refinement to iteratively update the binary sign matrices and the real-valued envelope factors.
Support for \mdbf is planned for a future release of \onecomp.

\subsection{Block-wise PTQ}
\label{sec:blockwise}

Within the broader framework of block-wise PTQ, \onecomp incorporates two complementary methods that operate at the level of an entire Transformer block.
Although both methods optimize block-level reconstruction objectives, they are introduced at different stages of the pipeline and serve distinct roles.
The first, block-wise compensation, is an optional procedure inserted during the layer-wise PTQ loop.
Its purpose is to update the remaining real number parameters in a partially quantized block, allowing it to absorb the errors introduced by already quantized sublayers.
The second, block-wise quantization, is a dedicated refinement stage applied after the layer-wise PTQ pass is complete.
Its purpose is to revisit a fully quantized block and jointly refine its parameters using a block-level distillation objective.
The crucial distinction lies not in the granularity of the methods, both of which operate over whole Transformer blocks, but in the parameters they update.
Specifically, block-wise compensation adjusts the remaining real number parameters before quantization, whereas block-wise quantization refines those that have already been quantized after the layer-wise PTQ procedure is complete.

\subsubsection{Block-wise Compensation}
\label{sec:blockwise_ft}

We first introduce \emph{block-wise compensation}, an optional procedure interleaved with sequential layer-wise PTQ.
Consider a Transformer block $b$ consisting of $R_b$ linear sublayers, which are quantized sequentially.
Suppose that the first $r-1$ sublayers have already been quantized, while the remaining sublayers $r, \ldots, R_b$ are still represented in floating point.
Before quantizing sublayer $r$, \onecomp temporarily optimizes the remaining FP16 parameters so that the resulting mixed-precision block more closely matches the original full-precision block.

Let $X_b$ denote the matrix of calibration inputs to block $b$.
We write $F_b^{(\mathrm{fp})}$ for the original full-precision operator of block $b$, and $F_{b,r}^{(\mathrm{mix})}(\cdot; \bm{\phi}_{b,r})$ for the mixed-precision block obtained after quantizing the first $r-1$ sublayers, where $\bm{\phi}_{b,r}$ denotes the FP16 parameters of the remaining sublayers.
Block-wise compensation minimizes the discrepancy between the outputs of the mixed-precision and full-precision blocks:
\begin{equation}
  \mathcal{L}_{b,r}^{\mathrm{comp}}
  =
  \frac{1}{n}
  \left\|
    F_{b,r}^{(\mathrm{mix})}(X_b; \bm{\phi}_{b,r})
    -
    F_b^{(\mathrm{fp})}(X_b)
  \right\|_F^2.
  \label{eq:bw_ft}
\end{equation}
During this optimization, the already quantized sublayers are kept fixed, and only the remaining FP16 parameters $\bm{\phi}_{b,r}$ are updated.
In our implementation, we optimize these parameters with Adam for 5 epochs using a learning rate of $10^{-4}$.
After this compensation step, sublayer $r$ is quantized using the chosen layer-wise PTQ method, and the same procedure may be repeated before quantizing each subsequent sublayer.

The goal of block-wise compensation is to enhance the sublayers within a block while they remain in FP16.
Once earlier sublayers in the same block have been quantized, their approximation errors perturb the hidden representations received by the remaining sublayers.
Block-wise compensation allows the remaining FP16 parameters to absorb part of this distortion before they are quantized.
It therefore serves as an intra-block adaptation mechanism embedded within the layer-wise PTQ pipeline, rather than as a post-quantization refinement stage applied after quantization is complete.
Importantly, this procedure is lightweight: it fits naturally within the standard single-GPU layer-wise PTQ workflow and requires no additional GPU resources.

\subsubsection{Block-wise Quantization}
\label{sec:blockwise_quantization}

We next describe block-wise quantization, which is applied only after the layer-wise PTQ pass has produced a fully quantized model.
At this point, every linear sublayer in every Transformer block has already been quantized.
The goal is no longer to modify floating-point parameters but to jointly refine the quantized parameters of an entire block.
By optimizing an entire block at once, this stage captures interactions among attention projections, MLP projections, residual connections, layer normalization, softmax, and activation functions—interactions that are only partially visible to purely layer-wise objectives.

\paragraph{Phase 1: Greedy Block-wise Distillation.}
Blocks are processed sequentially from input to output.
For each block $b$, \onecomp loads both the full-precision teacher block and the corresponding quantized student block onto the GPU.
Let $\{u_{b,j}^{(\mathrm{fp})}\}_{j=1}^{n}$ and $\{u_{b,j}^{(\mathrm{q})}\}_{j=1}^{n}$ denote the hidden states entering block $b$ in the full-precision teacher and quantized student models, respectively.
We define
\begin{equation}
Y_{b,j}^{(\mathrm{fp})}
=
F_b^{(\mathrm{fp})}\!\bigl(u_{b,j}^{(\mathrm{fp})}\bigr),
\qquad
Y_{b,j}^{(\mathrm{q})}(\bm{\theta}_b)
=
F_b^{(\mathrm{q})}\!\bigl(u_{b,j}^{(\mathrm{q})}; \bm{\theta}_b\bigr),
\end{equation}
where $F_b^{(\mathrm{q})}$ is the quantized block operator and $\bm{\theta}_b$ denotes the quantization variables associated with block $b$, including integer weights, their accompanying scale, zero-point, and clipping parameters.
The basic block-wise objective is then
\begin{equation}
  \min_{\bm{\theta}_b}
  \frac{1}{n}\sum_{j=1}^{n}
  \bigl\|
    Y_{b,j}^{(\mathrm{fp})}
    -
    Y_{b,j}^{(\mathrm{q})}(\bm{\theta}_b)
  \bigr\|_F^2.
  \label{eq:bw_obj}
\end{equation}

\onecomp can also enhance this reconstruction loss with a cosine-similarity term to preserve the directional structure of hidden representations:
\begin{equation}
  \mathcal{L}_{b}
  =
  \frac{1}{n}\sum_{j=1}^{n}
  \bigl\|
    Y_{b,j}^{(\mathrm{fp})}
    -
    Y_{b,j}^{(\mathrm{q})}
  \bigr\|_F^2
  +
  \lambda_{\cos}
  \cdot
  \frac{1}{n}\sum_{j=1}^{n}
  \Bigl(
    1 - \mathrm{CosSim}\bigl(
      Y_{b,j}^{(\mathrm{fp})},
      Y_{b,j}^{(\mathrm{q})}
    \bigr)
  \Bigr),
  \label{eq:bw_mixed}
\end{equation}
where $\mathrm{CosSim}(\bm{a}, \bm{b}) = \bm{a}^\top \bm{b} / (\|\bm{a}\| \|\bm{b}\|)$ is computed after flattening the hidden states, and $\lambda_{\cos} \ge 0$ controls the strength of the cosine term.

\paragraph{Phase 2: Cross-block Quantization.}
After greedy per-block distillation, \onecomp further reduces inter-block error accumulation by jointly refining adjacent block pairs.
Specifically, it applies a sliding window of size $K=2$ over neighboring blocks $(b, b+1)$, processed in the order $(0,1), (1,2), \ldots, (L-2, L-1)$, where $L$ is the total number of Transformer blocks.
For each pair, the optimization target is the output of the second block:
\begin{equation}
  \min_{\bm{\theta}_b, \bm{\theta}_{b+1}}
  \frac{1}{n}\sum_{j=1}^{n}
  \bigl\|
    Y_{b+1,j}^{(\mathrm{fp})}
    -
    Y_{b+1,j}^{(\mathrm{q})}(\bm{\theta}_b, \bm{\theta}_{b+1})
  \bigr\|_F^2.
  \label{eq:cbq}
\end{equation}
This cross-block stage explicitly models how quantization errors produced by block $b$ alter the input distribution and reconstruction quality of block $b+1$.
If the refined pairwise solution degrades either block relative to the result obtained after Phase 1, \onecomp rolls back to the earlier checkpoint.
This safeguard ensures that cross-block refinement is adopted only when it yields a genuine improvement.

\paragraph{Optimization through Smooth Straight-Through Estimator.}
Both Phase~1 and Phase~2 optimize quantized parameters through gradient descent, which requires differentiating through the discrete rounding operation.
The main technical challenge in block-wise quantization is that rounding is discrete and therefore has zero derivative almost everywhere.
To enable gradient-based optimization of quantized parameters, \onecomp uses a Smooth Straight-Through Estimator (Smooth STE).
The forward pass applies ordinary rounding, but the backward pass uses the differentiable surrogate
\begin{equation}
  \tilde{q}(x)
  =
  \lfloor x \rfloor
  +
  \sigma\bigl(k(\{x\} - 0.5)\bigr),
  \label{eq:smooth_ste}
\end{equation}
where $\{x\} = x - \lfloor x \rfloor$ is the fractional part of $x$, $\sigma$ is the sigmoid function, and $k > 0$ is a temperature parameter.
For small $k$, this surrogate is smooth and yields stable gradients; as $k$ increases, it approaches hard rounding.

To move gradually from smooth optimization to faithful discrete quantization, \onecomp employs \emph{Progressive STE Temperature Annealing} (PSTA), which linearly increases the temperature over training:
\begin{equation}
  k_e
  =
  k_{\min}
  +
  (k_{\max} - k_{\min})
  \cdot
  \frac{e}{E - 1},
  \label{eq:psta}
\end{equation}
where $e$ is the current epoch, $E$ is the total number of epochs, $k_{\min} = 2$, and $k_{\max} = 20$.
Early epochs emphasize smooth gradients and exploration, whereas later epochs emphasize near-discrete rounding and local refinement.
We use Adam with cosine learning-rate decay, 10\% linear warmup, and gradient clipping at 1.0.
Separate learning rates are assigned to different quantization variables: $10^{-4}$ for scales and zero-points, and $5 \times 10^{-5}$ for integer weights.

Taken together, these two methods provide complementary forms of block-level adaptation.
Block-wise compensation operates \emph{during} layer-wise PTQ by updating the remaining FP16 parameters inside a partially quantized block.
Block-wise quantization operates \emph{after} layer-wise PTQ by directly refining the quantized parameters of a fully quantized block.
This separation enables \onecomp to exploit block-level structure both before and after quantization is finalized.

\subsection{Global PTQ}
\label{sec:global}

A natural extension of layer-wise and block-wise refinement is to optimize the quantized model at the scale of the entire network.
This global PTQ stage is currently under development and planned for
a future release of \onecomp; we describe the design here for
completeness.
When sufficient GPU resources are available, this global PTQ stage can be framed as knowledge distillation from a full-precision teacher to a quantized student.
Unlike layer-wise and block-wise PTQ, which optimize local reconstruction objectives, global PTQ exposes the optimizer to the full computational graph, thereby enabling it to correct quantization errors that accumulate across many layers.
As a result, it is the most computationally demanding stage in the pipeline but also the most expressive.

\subsubsection{Loss Function}

Let $n$ denote the number of calibration token positions, $\mathcal{V}$ the vocabulary, and $L$ the number of Transformer layers.
For token position $j$, let $\bm{z}_j^{(\mathrm{fp})}, \bm{z}_j^{(\mathrm{q})} \in \mathbb{R}^{|\mathcal{V}|}$ be the teacher and student logits, respectively.
For each layer $l \in \{1,\dots,L\}$, let $\bm{h}_l^{(\mathrm{fp})}$ and $\bm{h}_l^{(\mathrm{q})}$ denote the corresponding teacher and student hidden states over the calibration set, flattened into vectors to compute cosine similarity. The overall objective combines three complementary terms:
\begin{equation}
  \mathcal{L}
  =
  \mathcal{L}_{\mathrm{KL}}
  +
  \lambda_{\mathrm{inter}} \mathcal{L}_{\mathrm{inter}}
  +
  \lambda_{\mathrm{ent}} \mathcal{L}_{\mathrm{ent}}.
  \label{eq:global_total}
\end{equation}
These terms serve distinct purposes.
The KL term aligns the final output distributions, the intermediate-state term preserves internal representations throughout the network, and the entropy term regularizes the student on the small calibration set.

\paragraph{Token-Level KL Distillation.}
The primary supervision is provided by KL distillation at a temperature $\tau$:
\begin{equation}
  \mathcal{L}_{\mathrm{KL}} =
  \frac{1}{n}
  \sum_{j=1}^{n}
  \sum_{v \in \mathcal{V}}
  p_{v,j}\bigl(\log p_{v,j} - \log q_{v,j}\bigr),~~
  p_{v,j} = \mathrm{softmax}\ab(\nicefrac{\bm{z}_j^{(\mathrm{fp})}}{\tau})_v,~~
  q_{v,j}
  = \mathrm{softmax}\ab(\nicefrac{\bm{z}_j^{(\mathrm{q})}}{\tau})_v .
  \label{eq:kl_loss}
\end{equation}
The default temperature is $\tau = 1$.
This objective encourages the quantized student to align with the teacher's output distribution rather than merely its most likely prediction, thereby preserving richer distributional information.

\paragraph{Intermediate Hidden-State Alignment.}
Matching only the final logits is often insufficient in low-bit settings because substantial distortion can occur in intermediate representations long before it appears at the output layer.
To mitigate this issue, we optionally align the hidden states at each layer:
\begin{equation}
  \mathcal{L}_{\mathrm{inter}} =
  \frac{1}{L} \sum_{l=1}^{L}\ab(1-\frac{\bm{h}_l^{(\mathrm{fp})\top}\bm{h}_l^{(\mathrm{q})}}{\|\bm{h}_l^{(\mathrm{fp})}\|_2\cdot\|\bm{h}_l^{(\mathrm{q})}\|_2}).
  \label{eq:inter_loss}
\end{equation}
This cosine-distance objective encourages the quantized student to preserve the direction of the teacher's internal representations across all layers.

\paragraph{Entropy Regularization.}
Since global PTQ is optimized on a relatively small calibration set, the student may become overconfident and overfit the observed examples.
We therefore incorporate the entropy regularizer
\begin{equation}
  \mathcal{L}_{\mathrm{ent}} = \frac{1}{n}
  \sum_{j=1}^{n} \sum_{v \in \mathcal{V}} q_{v,j}\log q_{v,j},
  \label{eq:ent_loss}
\end{equation}
Minimizing this term increases the entropy of the student distribution, improving regularization and enhancing robustness beyond the calibration set.

\subsubsection{Optimization Techniques}

Global PTQ is significantly more challenging to optimize than layer-wise or block-wise PTQ.
The parameter space encompasses the entire model, gradients navigate through numerous quantized operations, and the calibration set is limited.
Consequently, stable optimization benefits from several complementary techniques.

\paragraph{Sharpness-Aware Minimization.}
We first employ sharpness-aware minimization (SAM) \citep{foret2021sam}, which biases optimization towards flatter minima that are less sensitive to perturbations and, therefore, less prone to overfitting.
At each step, the parameters are perturbed in the local ascent direction $\bm{\epsilon} = \rho \cdot (\nicefrac{\nabla_{\bm{\theta}} \mathcal{L}}{\|\nabla_{\bm{\theta}} \mathcal{L}\|_2})$,
and the actual update is calculated at the perturbed point:
\begin{equation}
  \bm{\theta} \leftarrow \bm{\theta} - \eta
  \nabla_{\bm{\theta}} \mathcal{L}(\bm{\theta} + \bm{\epsilon}).
  \label{eq:sam}
\end{equation}
In practice, SAM avoids narrow optima that fit the calibration data well but generalize poorly.

\paragraph{Progressive Layer Unfreezing.}

Optimizing all layers simultaneously from the start can lead to
instability, especially when the student experiences significant
quantization noise.
We therefore use progressive layer unfreezing, starting with the layers closest to the output and gradually extending optimization to earlier layers.
At epoch $e$, only layers satisfying $l \ge L - (e+1)\cdot \left\lceil \nicefrac{L}{E} \right\rceil$
are trainable, where $E$ is the total number of epochs.
This schedule improves stability by first adapting the layers that most directly affect the logits and only later allowing optimization to reshape deeper internal computations.

Overall, global PTQ offers the most comprehensive form of post-training refinement in the pipeline.
While layer-wise and block-wise methods correct local reconstruction errors, global PTQ directly optimizes the end-to-end behavior of the entire quantized model under full-model supervision.
Although this stage is significantly more expensive, it provides a principled approach to recovering long-range interactions and cross-layer dependencies that are difficult to capture with local objectives alone.
We plan to integrate this global PTQ stage into \onecomp in a future release.

\subsection{Quantization-Aware Fine-tuning}
\label{sec:finetuning}

As a final optional stage, \onecomp supports quantization-aware fine-tuning via QLoRA~\citep{dettmers2023qlora}.
At this stage, the quantized backbone remains frozen, and adaptation is achieved by training only a small set of low-rank adapter parameters.
Specifically, for each linear layer with a quantized weight matrix $\widehat{W} \in \mathbb{R}^{N \times M}$, we introduce two trainable low-rank matrices,
$B \in \mathbb{R}^{N \times r}$ and $A \in \mathbb{R}^{r \times M}$, where $r \ll \min(N,M)$.
The resulting effective weight matrix is
\begin{equation}
  W_{\mathrm{eff}} = \widehat{W} + BA.
  \label{eq:qlora}
\end{equation}
During fine-tuning, only the adapter matrices $A$ and $B$ are updated, while the quantized backbone weight $\widehat{W}$ remains fixed.
Consequently, the number of trainable parameters per layer is only $(N+M)r$, typically a small fraction of the $NM$ parameters in the corresponding full-rank matrix.

This stage serves two complementary purposes.
It facilitates domain adaptation.
When the quantized model is deployed on data that differ from the distribution observed during pretraining, the low-rank adapters can learn task- or domain-specific corrections without altering the compressed backbone.
Furthermore, it enables recovery from quantization errors. Even with minimal or no domain shift, quantization can introduce residual distortions that post-training quantization may not fully eliminate.
The low-rank correction $BA$ can absorb some of this residual error, thereby recovering part of the accuracy lost during quantization.

\onecomp supports this stage with user-specified training datasets and standard fine-tuning hyperparameters, allowing users to flexibly trade off adaptation quality, computational cost, and parameter efficiency based on the requirements of the target deployment setting.
\section{Experimental Evaluation}
\label{sec:experiments}

We evaluate the \onecomp pipeline through targeted experiments designed to test its core design hypotheses.
Our empirical study is organized around five questions:
\begin{enumerate}
    \item Does AutoBit yield better results than uniform assignment? (Section~\ref{sec:exp_autobit})
    \item Do QEP and LPCD enhance layer-wise post-training quantization? (Section~\ref{sec:exp_layerwise})
    \item Does JointQ improve the GPTQ parameterization? (Section~\ref{sec:exp_jointq})
    \item Can structured binary factors maintain model quality in the extreme low-bit regime? (Section~\ref{sec:exp_dbf})
    \item Does performance improve monotonically with increased computational resources? (Section~\ref{sec:exp_progress})
\end{enumerate}
Each subsection addresses a question, presents empirical evidence, and highlights practical implications.

\subsection{Setup}
\label{subsec:setup}

We conduct an empirical study on a representative set of open-weight decoder-only LLMs, which includes two model families and various architectural scales: Llama3 8B \citep{grattafiori2024llama} and Qwen3-8B and 14B \citep{yang2025qwen3}.
These models represent a range of modern transformer configurations used in practice, including dense decoder architectures and grouped-query attention variants.
As evaluation metrics, we report perplexity on WikiText-2 as the primary measure of language modeling fidelity and zero-shot accuracy on a standard downstream benchmark suite implemented with \texttt{lm-eval-harness}, which includes ARC-Challenge, ARC-Easy, PIQA, and WinoGrande.
Following common post-training quantization practices, we evaluate both the original full-precision model and its quantized counterparts under the same protocol.
Lower perplexity indicates better preservation of generative quality, while higher zero-shot accuracy indicates improved task performance.
For calibration, we use samples drawn from C4 with a maximum sequence length of 2048 tokens.
Unless otherwise stated, the calibration set is shared across methods in each comparison to isolate the effect of the quantization algorithm.
Detailed choices of bit-width, grouping strategy, and other method-specific hyperparameters are deferred to the corresponding experimental subsections, where each component is analyzed under its appropriate comparison setting.

\subsection{Mixed-Precision Allocation} \label{sec:exp_autobit}

We evaluate \autobit, which performs module-wise mixed-precision allocation under a fixed average bit-per-weight (bpw) budget.
Unlike conventional uniform quantization, which applies the same quantization configuration to all modules, \autobit selects different quantization settings for different modules according to their estimated sensitivity.

We consider mixed-precision allocation over individual Transformer linear modules, including attention projections
(\texttt{q\_proj}, \texttt{k\_proj}, \texttt{v\_proj}, \texttt{o\_proj})
and MLP projections
(\texttt{gate\_proj}, \texttt{up\_proj}, \texttt{down\_proj}).
For each module, \autobit selects a quantization configuration from a candidate set defined by bit-width and group size.
In our experiments, the bit-width candidates are $\{2, 3, 4, 5, 6, 7, 8\}$ and the group sizes of $\{128, \texttt{per-channel}\}$, where $\texttt{per-channel}$ denotes channel-wise grouping.
We evaluate two target budgets expressed as average bits per weight
(bpw): $4.16$ and $4.00$, corresponding to uniform GPTQ with 4-bit
group-128 and 4-bit channel-wise quantization, respectively.

We compare three allocation strategies under the same target budget:
(1) \textbf{Uniform}, which applies a single quantization configuration to all eligible modules;
(2) \textbf{Naive}, which performs mixed-precision allocation without activation information;
and
(3) \textbf{Activation-aware (diag.\ only)}, which augments the allocation score with diagonal activation statistics.
All methods use the same calibration/evaluation pipeline; the only difference is how the module-wise quantization budget is allocated.

\begin{figure}[htbp]
    \centering
    \includegraphics[width=\linewidth]{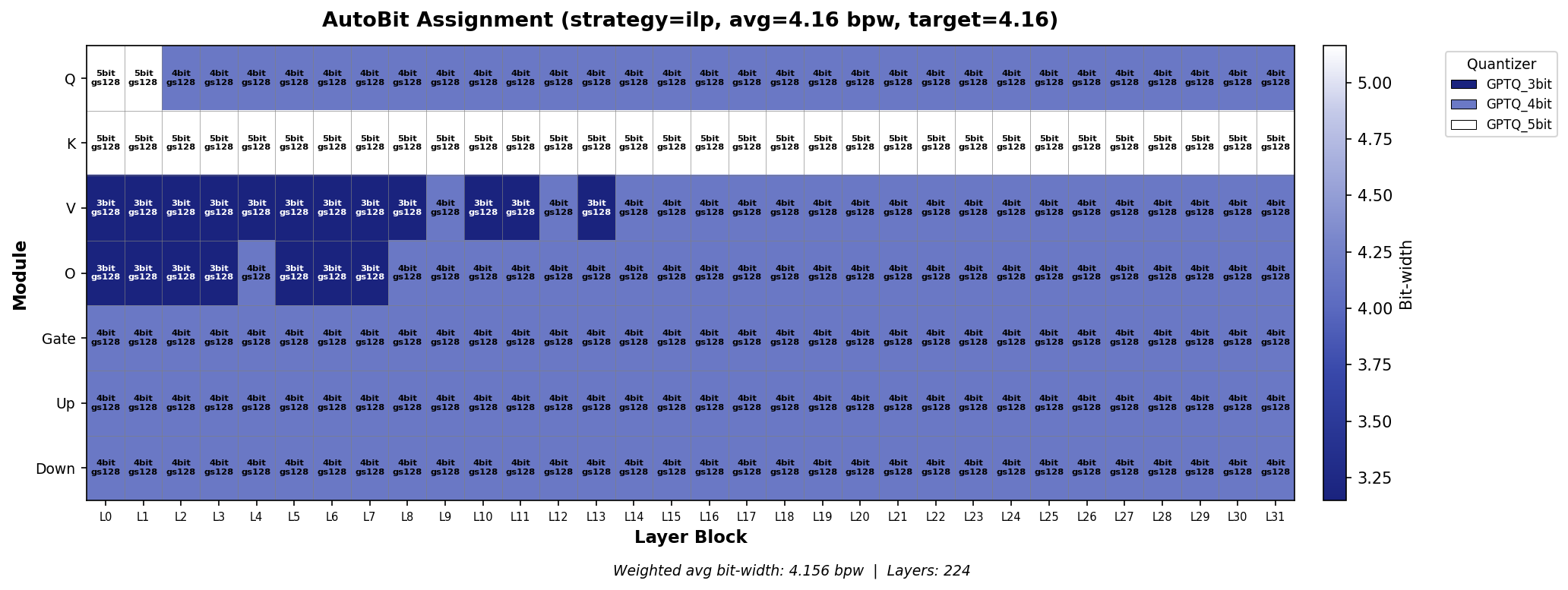}
    \includegraphics[width=\linewidth]{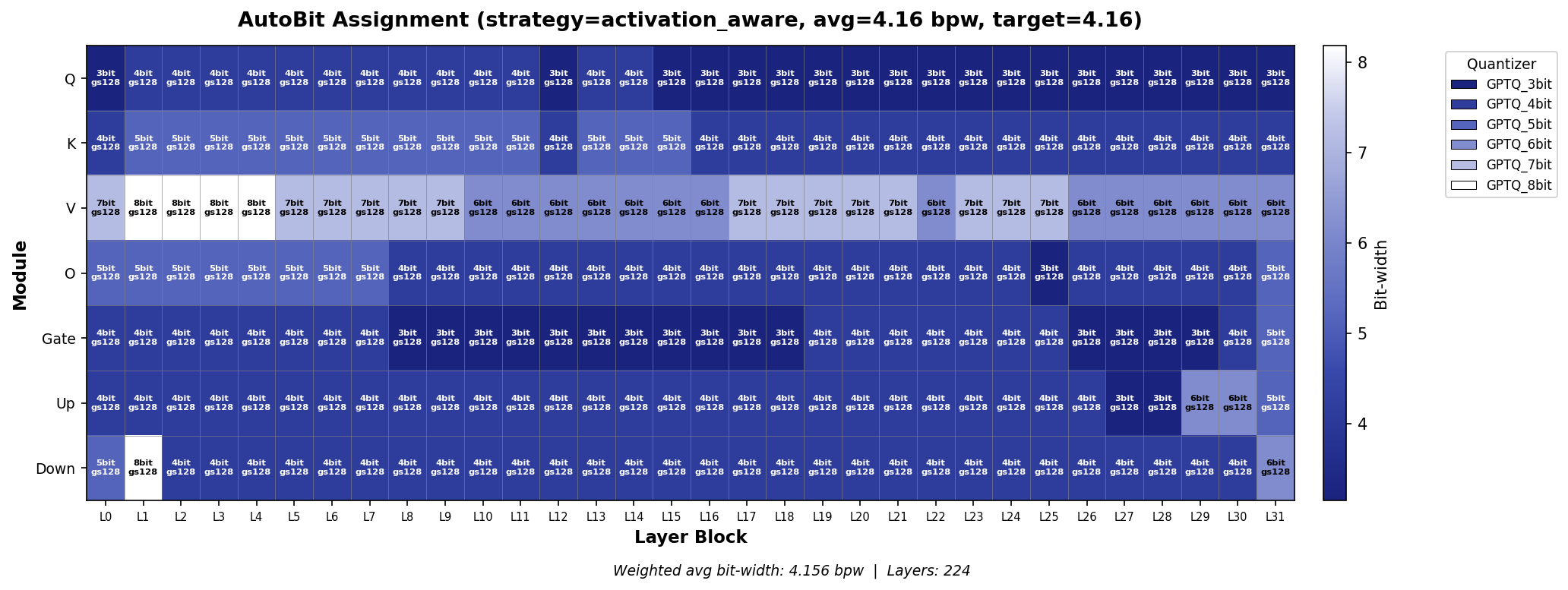}
    \caption{Example module-wise bit allocation on Llama~3 under a 4.16 average bpw budget, which is equivalent to 4-bit quantization with a group size of 128.
    Top: naive mixed-precision allocation.
    Bottom: activation-aware allocation using diagonal activation statistics only.
    The activation-aware variant tends to reserve higher precision for modules with stronger activation outliers or higher activation sensitivity.}
    \label{fig:autobit_alloc}
\end{figure}

\begin{table*}[htbp]
\centering
\small
\setlength{\tabcolsep}{3.0pt}
\renewcommand{\arraystretch}{1.12}
\caption{
AutoBit mixed-precision allocation under fixed average bit-per-weight (bpw) budgets.
For each model, the full-precision (FP) reference is shown once and shared across all quantized variants.
We compare three allocation methods: Uniform, Naive, and \textsc{Act.-aware} (diagonal activation only).
For each bpw budget, perplexity (PPL; lower is better) and average 0-shot accuracy (ACC; higher is better) are reported separately.
}
\label{tab:autobit_combined}
\begin{threeparttable}
\begin{tabular}{lcccccccc}
\toprule
\multicolumn{1}{c}{\multirow{3}{*}{Model}}
& \multicolumn{2}{c}{FP}
& \multicolumn{1}{c}{\multirow{3}{*}{Method}}
& \multicolumn{4}{c}{Average bit per weight (bpw)} \\
\cmidrule(lr){2-3}
\cmidrule(lr){5-8}
& \multirow{2}{*}{PPL} & \multirow{2}{*}{ACC}
& & \multicolumn{2}{c}{4.16 \ (4b/128)}
  & \multicolumn{2}{c}{4.00 \ (4b/ch)} \\
\cmidrule(lr){5-6}
\cmidrule(lr){7-8}
& & & & PPL & ACC & PPL & ACC \\
\midrule

\multirow{3}{*}{Llama-3-8B}
& \multirow{3}{*}{6.14}
& \multirow{3}{*}{0.715}
& Uniform
& 9.53 & 0.697 & 383.26 & 0.595 \\
& & & Naive
& 10.28 & 0.691 & 10.95 & 0.682 \\
& & & \textsc{Act.-aware}
& \textbf{6.85} & \textbf{0.702} & \textbf{7.04} & \textbf{0.701} \\
\midrule

\multirow{3}{*}{Qwen3-8B}
& \multirow{3}{*}{9.72}
& \multirow{3}{*}{0.707}
& Uniform
& 10.92 & \textbf{0.706} & 10.65 & 0.679 \\
& & & Naive
& 10.95 & 0.705 & 11.11 & \textbf{0.702} \\
& & & \textsc{Act.-aware}
& \textbf{10.25} & 0.700 & \textbf{10.56} & 0.693\\

\bottomrule
\end{tabular}

\end{threeparttable}
\end{table*}

Table~\ref{tab:autobit_combined} summarizes the main results on Llama~3 and Qwen3.
Figure~\ref{fig:autobit_alloc} visualizes an example module-wise allocation pattern for Llama~3 under the naive and activation-aware variants.

Table~\ref{tab:autobit_combined} reveals two distinct performance regimes.
On Llama-3-8B, the benefit of mixed-precision allocation is substantial.
At 4.16 average bpw, the activation-aware variant achieves a perplexity of 6.85, close to the full-precision baseline of 6.14, whereas uniform assignment degrades to 9.53 and the naive variant to 10.28.
The gap widens further at 4.00 bpw (channel-wise grouping), where uniform quantization collapses entirely, while the activation-aware method maintains a perplexity of 7.04 and accuracy of 0.701, nearly matching the full-precision reference.
This indicates that channel-wise uniform 4-bit quantization is catastrophically fragile for Llama-3-8B: a small number of highly sensitive modules dominate the overall error, and activation-aware mixed-precision allocation effectively avoids this failure mode by directing higher precision to those modules.

On Qwen3-8B, the activation-aware variant consistently achieves the lowest perplexity (10.25 at 4.16~bpw; 10.56 at 4.00~bpw), while accuracy differences among the three strategies remain within one percentage point.
This suggests that Qwen3-8B exhibits less inter-module sensitivity heterogeneity than Llama-3-8B, making the uniform baseline already a reasonable allocation and limiting the headroom for mixed-precision gains.

Compared to the naive mixed-precision strategy, the activation-aware variant further improves allocation by considering activation statistics.
Modules with stronger activation outliers or higher sensitivity tend to receive more bits, leading to a more effective use of the limited precision budget.
For example, Act.-aware assigns a higher bit-width to the down module in Layer 1, which exhibits large activation outliers.
Besides, the naive approach does not assign a high bit-width to the module. (See Figure~\ref{fig:autobit_alloc})

Overall, \autobit's mixed-precision allocation is the recommended strategy regardless of the model architecture.
This indicates that a uniform bit-width leaves substantial performance on the table, even when the total memory cost is fixed.

\subsection{Layer-wise PTQ with Error-Propagation Correction}
\label{sec:exp_layerwise}

The layer-wise regime constitutes the most relevant operating point of \onecomp, as it remains feasible within a strict single-GPU memory budget. We therefore focus this experiment on a key deployment question: can explicit correction substantially improve layer-wise post-training quantization (PTQ) without relaxing the underlying hardware constraints?
To address this question, we compare three progressively robust correction strategies based on the same GPTQ base quantizer. The first, Base, corresponds to standard layer-wise GPTQ without any cross-layer correction. The second, \qep, modifies the quantization target of each linear layer using the error-propagation matrix accumulated from all previously quantized layers, as described in Section~\ref{sec:qep}. The third, Submodule, corresponds to LPCD and extends \qep by additionally accounting for the error propagated through the residual stream of each transformer block.

In a standard transformer block, the outputs of the attention sub-layer and the MLP sub-layer are each added to the residual stream through skip connections. The submodule treats these residual-connected units as coherent submodules and introduces an auxiliary cross-term that captures the discrepancy between full-precision and quantized residual activations. This residual cross-term provides a second-order correction to the weight target, complementing the standard layer-level \qep adjustment. For each layer whose output directly feeds into a residual connection, such as the output or down projection, the corrected target includes an additional term proportional to the inverse Hessian applied to the residual-stream error. In this way, the formulation preserves the single-GPU, layer-sequential workflow of \qep while accounting for the coupling between quantization error and the residual stream, which standard layer-wise objectives ignore.
Table~\ref{tab:layerwise_combined} reports perplexity and average zero-shot accuracy for Llama-3-8B under 4-bit and 3-bit quantization, using a group size of 128 with per-channel grouping. We present the full-precision reference once, then compare the three correction strategies under each quantization setting.

\begin{table}[tb]
\centering
\small
\setlength{\tabcolsep}{3.0pt}
\renewcommand{\arraystretch}{1.12}
\caption{
Layer-wise PTQ on Llama-3-8B under a single-GPU budget.
The full-precision (FP) reference is shown once.
Base: GPTQ without correction.
\qep: layer-level error-propagation correction.
Submodule: \qep extended with residual-stream correction.
PPL: WikiText-2 perplexity ($\downarrow$).
ACC: average 0-shot accuracy ($\uparrow$).
}
\label{tab:layerwise_combined}
\begin{tabular}{lccccccccccc}
\toprule
\multicolumn{1}{c}{\multirow{3}{*}{Model}}
& \multicolumn{2}{c}{FP}
& \multicolumn{1}{c}{\multirow{3}{*}{Method}}
& \multicolumn{8}{c}{Quantization setting} \\
\cmidrule(lr){2-3}
\cmidrule(lr){5-12}
& \multirow{2}{*}{PPL} & \multirow{2}{*}{ACC}
& & \multicolumn{2}{c}{4b/128}
  & \multicolumn{2}{c}{4b/ch}
  & \multicolumn{2}{c}{3b/128}
  & \multicolumn{2}{c}{3b/ch} \\
\cmidrule(lr){5-6}
\cmidrule(lr){7-8}
\cmidrule(lr){9-10}
\cmidrule(lr){11-12}
& & & & PPL & ACC & PPL & ACC & PPL & ACC & PPL & ACC \\
\midrule

\multirow{3}{*}{Llama-3-8B}
& \multirow{3}{*}{6.14}
& \multirow{3}{*}{0.715}
& Base
& 12.66 & 0.697
& 665.94 & 0.533
& 45.22 & 0.521
& 1721.06 & 0.386 \\
& & & \qep
& 6.66 & \textbf{0.710}
& 7.67 & 0.688
& 8.95 & 0.634
& \textbf{17.93} & \textbf{0.496} \\
& & & Submodule
& \textbf{6.59} & 0.709
& \textbf{7.33} & \textbf{0.703}
& \textbf{8.54} & \textbf{0.640}
& 22.40 & 0.482 \\
\bottomrule
\end{tabular}
\end{table}

First, Base GPTQ without correction severely degrades performance, particularly under per-channel quantization, where perplexity rises dramatically and downstream accuracy drops to near-random levels. This indicates that naive layer-wise PTQ is insufficient for aggressive quantization in this architecture.
Second, \qep consistently restores model quality across all settings. It keeps perplexity close to the full-precision baseline and substantially recovers downstream accuracy, showing that error-propagation correction effectively compensates for quantization errors overlooked by standard layer-wise objectives.
Third, the Submodule offers additional gains over \qep in moderately aggressive settings. In the 4-bit settings and the 3-bit group-size-128 setting, it achieves lower perplexity and comparable or better accuracy, suggesting that modeling residual-stream coupling provides benefits beyond layer-level correction.

\subsection{\jointq versus Standard GPTQ} \label{sec:exp_jointq}

Both GPTQ and \jointq represent quantized weights in the same standard packed-weight format, utilizing symmetric quantization with bit-widths of $\{3,4\}$ and group sizes of $\{128, \texttt{per-channel}\}$. For both methods, we use the default configuration, with activation ordering and MSE-based grid tuning disabled. For \jointq, the regularization strength is set to $\lambda=0.2$.
Table~\ref{tab:jointq_combined} reports perplexity and average zero-shot accuracy for GPTQ and \jointq on Llama-3-8B, Qwen3-8B, and Qwen3-14B under each quantization setting. Boldface indicates the better result between the two methods for each metric and setting.

\begin{table*}[htbp]
\centering
\small
\setlength{\tabcolsep}{3.0pt}
\renewcommand{\arraystretch}{1.12}
\caption{
Comparison between \jointq and standard GPTQ under the same packed quantized-weight format, with activation ordering and MSE-based tuning disabled.
For each model, the full-precision (FP) reference is shown once.
Here, $128$ denotes group-wise quantization with group size $128$, and \texttt{ch} denotes per-channel quantization.
We report perplexity (PPL $\downarrow$) and average zero-shot accuracy (ACC $\uparrow$).
Boldface indicates the better result between GPTQ and \jointq for each setting.
}
\label{tab:jointq_combined}
\begin{threeparttable}
\begin{tabular}{lccccccccccc}
\toprule
\multicolumn{1}{c}{\multirow{3}{*}{Model}}
& \multicolumn{2}{c}{FP}
& \multicolumn{1}{c}{\multirow{3}{*}{Method}}
& \multicolumn{8}{c}{Quantization setting} \\
\cmidrule(lr){2-3}
\cmidrule(lr){5-12}
& \multirow{2}{*}{PPL} & \multirow{2}{*}{ACC}
& & \multicolumn{2}{c}{4b/128}
  & \multicolumn{2}{c}{4b/ch}
  & \multicolumn{2}{c}{3b/128}
  & \multicolumn{2}{c}{3b/ch} \\
\cmidrule(lr){5-6}
\cmidrule(lr){7-8}
\cmidrule(lr){9-10}
\cmidrule(lr){11-12}
& & & & PPL & ACC & PPL & ACC & PPL & ACC & PPL & ACC \\
\midrule

\multirow{2}{*}{Llama-3-8B}
& \multirow{2}{*}{6.14}
& \multirow{2}{*}{0.715}
& GPTQ
& 12.66 & 0.697
& 665.94 & 0.533
& 45.22 & 0.521
& 1721.06 & 0.386 \\
& & & \jointq
& \textbf{6.67} & \textbf{0.705}
& \textbf{8.46} & \textbf{0.676}
& \textbf{9.26} & \textbf{0.650}
& \textbf{21.15} & \textbf{0.583} \\
\midrule

\multirow{2}{*}{Qwen3-8B}
& \multirow{2}{*}{9.72}
& \multirow{2}{*}{0.707}
& GPTQ
& 10.29 & \textbf{0.697}
& \textbf{10.97} & \textbf{0.672}
& \textbf{11.71} & 0.654
& \textbf{20.21} & 0.496 \\
& & & \jointq
& 10.29 & 0.695
& 11.33 & 0.660
& 12.38 & \textbf{0.668}
& 42.10 & \textbf{0.506} \\
\midrule

\multirow{2}{*}{Qwen3-14B}
& \multirow{2}{*}{8.64}
& \multirow{2}{*}{0.739}
& GPTQ
& \textbf{8.85} & 0.736
& \textbf{9.15} & 0.720
& 10.10 & 0.698
& \textbf{13.72} & 0.581 \\
& & & \jointq
& 8.88 & \textbf{0.739}
& 9.96 & \textbf{0.723}
& \textbf{9.99} & \textbf{0.701}
& 23.55 & \textbf{0.651} \\
\bottomrule
\end{tabular}

\end{threeparttable}
\end{table*}

On Llama-3-8B, \jointq consistently outperforms GPTQ in both perplexity and accuracy across all quantization settings. The gains are especially pronounced under per-channel quantization, where standard GPTQ shows significant degradation. In the Qwen3 models, the comparison is more nuanced: \jointq generally yields stronger downstream accuracy, particularly under more aggressive settings like 3b/ch, while GPTQ often achieves lower perplexity. This discrepancy between perplexity and accuracy suggests that the two methods preserve different aspects of model behavior. Nonetheless, the consistent accuracy improvements obtained by \jointq indicate that jointly optimizing scales and integer weights better maintains task-relevant structure than the standard alternating procedure.

\subsection{Extreme Low-Bit Quantization} \label{sec:exp_dbf}

As foundation models continue to scale, the need for extreme compression is becoming more urgent. Practical scenarios such as on-device inference, edge deployment, and memory-constrained serving require bit-widths significantly below the conventional 3–4 bit range. This experiment examines whether structured binary-factor representations can preserve useful model quality at 1.00 and 1.50 bits per weight (BPW), a regime where standard uniform quantization typically fails.

We compare DBF~\citep{boza2026addition} and \mdbf~\citep{ichikawa2025mdbf} across five models ranging from 0.6B to 8B parameters. 
Since smaller models have less redundancy in their weight distributions, they pose a greater challenge for extreme low-bit compression; therefore, we include sub-billion-scale models to test the methods under the most demanding conditions.
For \mdbf, we set the envelope rank to $l = 3$ for LLaMA-2-7B and $l = 4$ for all other models. Both methods are evaluated using the complete three-stage \onecomp pipeline: (i) layer-wise PTQ with MSVID and ADMM optimization (1,000 outer iterations and 3 inner iterations), (ii) block-wise PTQ with continuous-parameter refinement (learning rate $10^{-3}$ for 4 epochs), and (iii) global PTQ with KL distillation from the full-precision teacher (learning rate $5 \times 10^{-5}$ for 3 epochs). For calibration, we utilize 512 samples from WikiText-2, with a maximum sequence length of 2,048 tokens at all three stages.
Table~\ref{tab:lowbit} reports the final results after the complete pipeline, including WikiText-2 perplexity (PPL; lower is better) and average zero-shot accuracy (ACC; higher is better) over ARC-Challenge, ARC-Easy, PIQA, and WinoGrande. 

\begin{table*}[t]
\centering
\small
\setlength{\tabcolsep}{3.0pt}
\renewcommand{\arraystretch}{1.12}
\caption{
Extreme low-bit quantization results after the full three-stage pipeline.
For each bit budget we report WikiText-2 perplexity (PPL$\downarrow$) and average zero-shot accuracy (ACC$\uparrow$).
Boldface indicates the better result between DBF and \mdbf.
}
\label{tab:lowbit}
\begin{tabular}{lcccccccc}
\toprule
\multicolumn{1}{c}{\multirow{3}{*}{Model}}
& \multicolumn{4}{c}{1.00 BPW}
& \multicolumn{4}{c}{1.50 BPW} \\
\cmidrule(lr){2-5}
\cmidrule(lr){6-9}
& \multicolumn{2}{c}{DBF} & \multicolumn{2}{c}{\mdbf}
& \multicolumn{2}{c}{DBF} & \multicolumn{2}{c}{\mdbf} \\
\cmidrule(lr){2-3}
\cmidrule(lr){4-5}
\cmidrule(lr){6-7}
\cmidrule(lr){8-9}
& PPL & ACC & PPL & ACC
& PPL & ACC & PPL & ACC \\
\midrule
Qwen3-0.6B     & 43.07 & 0.412 & \textbf{34.45} & \textbf{0.426} & 30.80 & 0.435 & \textbf{26.56} & \textbf{0.446} \\
Llama-3.2-1B   & 21.48 & 0.465 & \textbf{18.29} & \textbf{0.473} & 16.58 & 0.491 & \textbf{15.00} & \textbf{0.492} \\
TinyLlama-1.1B & 13.57 & 0.449 & \textbf{12.29} & \textbf{0.449} & 10.38 & \textbf{0.491} & \textbf{9.84} & 0.490 \\
Llama-2-7B     & 9.95  & 0.480 & \textbf{9.36}  & \textbf{0.489} & 7.34  & 0.548 & \textbf{7.17}  & \textbf{0.564} \\
Llama-3-8B     & 14.76 & 0.505 & \textbf{13.25} & \textbf{0.518} & 10.61 & 0.560 & \textbf{10.03} & \textbf{0.572} \\
\bottomrule
\end{tabular}
\end{table*}

Table~\ref{tab:lowbit} shows that \mdbf consistently achieves lower perplexity than DBF across all five models and both bit budgets. In every setting, \mdbf achieves lower perplexity, indicating that its low-rank magnitude envelope is more expressive than the rank-one diagonal scaling used in DBF. This advantage increases at 1.00 BPW and for larger models, suggesting that \mdbf better exploits the greater redundancy present in larger weight matrices. By contrast, at 1.50 BPW, the two methods perform more similarly for smaller models, where the recoverable structure is more limited.
This advantage in perplexity is also reflected in downstream performance. \mdbf generally achieves higher average zero-shot accuracy, with the most significant gains observed in larger models at 1.00 BPW. Although a substantial gap from full precision remains at the most aggressive setting, this gap narrows considerably at 1.50 BPW, particularly for larger models. Overall, these results indicate that structured binary-factor quantization can preserve a meaningful fraction of model capability at extremely low bit-widths and that \mdbf provides a clear improvement over DBF in this respect.
Support for \mdbf is planned for the next release of \onecomp.

\subsection{Progressive Quality Improvement Across Execution Regimes}
\label{sec:exp_progress}

\begin{figure}[tb]
  \centering
  \includegraphics[width=\linewidth]{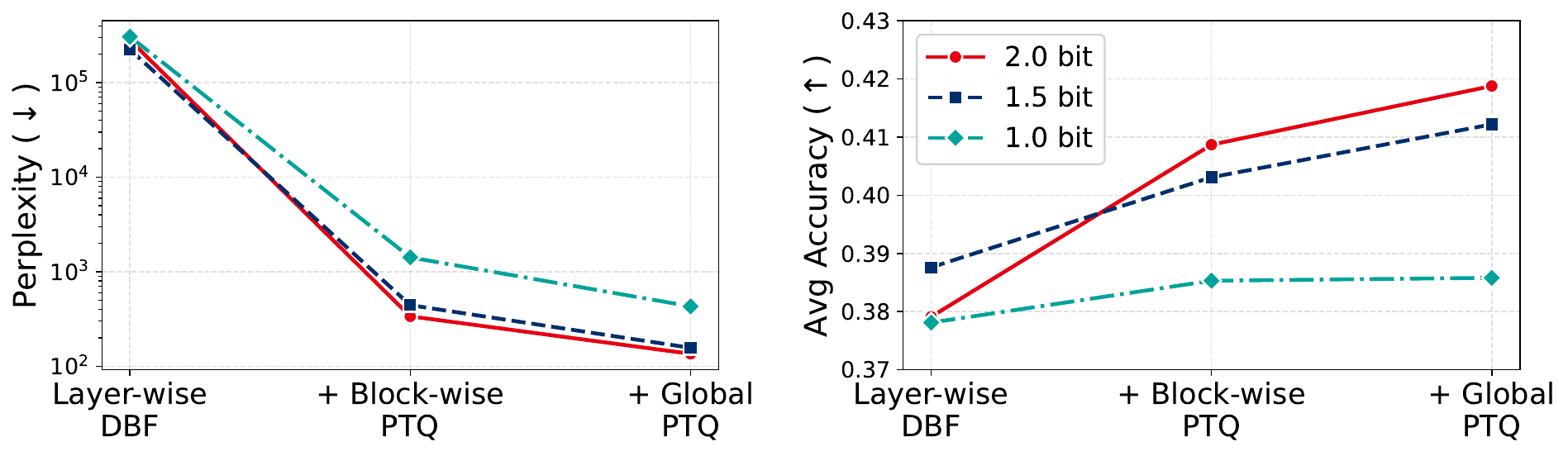}
  \caption{Progressive quality improvement on Llama-3.2-1B under DBF quantization at 2.0, 1.5, and 1.0 bits per weight.
  Each pipeline stage, layer-wise, block-wise, and global PTQ, provides monotonic accuracy improvement regardless of bit-width.}
  \label{fig:progressive}
\end{figure}

A core premise of \onecomp is that model quality improves monotonically as more compute is invested, with each execution regime refining the result of the previous one. We evaluate this hypothesis in the extreme low-bit setting by applying DBF quantization at 1.0, 1.5, and 2.0 bits per weight, progressively enabling three stages: layer-wise PTQ, which performs independent per-layer ADMM decomposition; block-wise PTQ, which applies per-block MSE distillation from an FP16 teacher with cross-block refinement; and global PTQ, which conducts end-to-end KL distillation over the full model. In the block-wise stage, DBF scaling vectors and binary matrices are optimized using SmoothSign STE with progressive temperature annealing, while the global stage utilizes Adam with Lookahead and entropy regularization.
Figure~\ref{fig:progressive} reports the average zero-shot accuracy of Llama-3.2-1B on the four benchmark tasks described in Section~\ref{subsec:setup}. Each bar group corresponds to one stage of the pipeline, the colors indicate the target bit-width, and the dashed line marks the FP16 baseline.

Two clear patterns emerge. First, each additional stage yields a monotonic improvement in accuracy across all three bit-widths: block-wise PTQ consistently improves upon the layer-wise baseline, and global PTQ provides further gains. These improvements are larger at higher bit-widths, where the scaling parameters allow for greater flexibility in refinement. In contrast, at 1 bit, the gains are smaller because performance is more strongly constrained by the discrete representation.
Second, global PTQ remains beneficial even in the most aggressive settings. At all bit-widths, it further reduces the gap between the layer-wise result and the FP16 baseline, indicating that the end-to-end KL objective captures cross-layer dependencies not addressed by block-level MSE alone.
\section{Related Work}
\label{sec:related}

Post-training quantization for large language models is one of the most active research areas in efficient deep learning.
This section surveys the landscape organized by the stages of the quantization pipeline, providing context for \onecomp's design choices.

\paragraph{Surveys and Benchmarks.}
Several comprehensive surveys cover the rapidly evolving quantization landscape.
\citet{gong2024survey} provides a systematic overview of low-bit LLM quantization, covering the fundamentals, systems, and algorithms.
\citet{gholami2022survey} offers a broader perspective on quantization techniques in deep learning.
\citet{zhao2025benchmarking} presents a unified benchmark that decouples quantization methods into pre-quantization transformations and error mitigation, enabling fair comparisons among methods.
These surveys highlight the critical importance of algorithm selection and workflow design, which motivates \onecomp's automated pipeline.

\paragraph{Investigation and Mixed-Precision Bit Allocation.}
The observation that different layers exhibit varying quantization sensitivities has motivated mixed-precision quantization.
APTQ~\citep{guan2024aptq} formulates bit allocation as an attention-aware optimization, using the Hessian trace as a sensitivity metric.
SqueezeLLM~\citep{kim2024squeezellm} employs a dense-and-sparse decomposition that performs heterogeneous allocation by isolating outlier weights into a sparse representation while aggressively quantizing the dense component.
\citet{dettmers2023case} studies scaling laws for 4-bit inference and demonstrates that mixed-precision consistently outperforms uniform allocation.
\onecomp's \autobit planner addresses an integer linear programming problem using SCIP.

\paragraph{Preprocessing and Outlier Mitigation.}
Outlier channels in transformer weights and activations pose a fundamental challenge for low-bit quantization.
SmoothQuant~\citep{xiao2023smoothquant} migrates quantization difficulty from activations to weights using per-channel scaling, enabling accurate W8A8 quantization.
QuIP~\citep{chee2024quip} applies random orthogonal transforms to uniformly distribute outlier energy, while QuIP\#~\citep{tseng2024quip} replaces random rotations with Hadamard transforms and incorporates lattice codebooks.
QuaRot~\citep{ashkboos2024quarot} extends rotation into an end-to-end framework that includes hidden states, attention matrices, and KV caches.
SpinQuant~\citep{liu2024spinquant} demonstrates that learned Cayley-parameterized rotations significantly outperform random rotations.
OmniQuant~\citep{shao2023omniquant} jointly learns per-channel clipping bounds and equivalence transforms within a differentiable framework.

\paragraph{Layer-wise PTQ.}
Layer-wise PTQ is the most commonly used approach, because it processes one linear layer at a time on a single GPU.
GPTQ~\citep{frantar2022gptq} pioneered Hessian-aware column-by-column rounding using the Optimal Brain Surgeon framework.
AWQ~\citep{lin2024awq} proposed activation-aware channel scaling that avoids explicit Hessian computation.
ZeroQuant~\citep{yao2022zeroquant} combines weight and activation quantization with efficient inference kernels.
QEP~\citep{arai2025quantization} addresses the error propagation issue by modifying the quantization target to compensate for upstream errors through a closed-form correction.
LPCD~\citep{ichikawa2025lpcd} generalizes this to arbitrary submodules using a relax-then-project coordinate descent framework.
\jointq further improves GPTQ-format models by jointly optimizing scales and integers.

\paragraph{Block-wise PTQ.}
Block-wise methods optimize all quantization parameters within a Transformer block.
BRECQ~\citep{li2021brecq} pioneered block-wise reconstruction, demonstrating significant error reduction compared to layer-wise approaches and advancing post-training quantization (PTQ) to INT2 for the first time.
CBQ~\citep{ding2023cbq} extends this concept to the cross-block optimization of adjacent block pairs.
OmniQuant~\citep{shao2023omniquant} supports learnable block-wise parameters.

\paragraph{Global PTQ and Quantization-aware Training.}
Global methods consider all layers simultaneously, providing high quality at a considerable resource cost.
LLM-QAT~\citep{liu2023llmqat} applies data-free quantization-aware training using synthetic data generated by the model itself.
PV-Tuning~\citep{malinovskii2024pv} goes beyond straight-through estimation for extreme compression by optimizing in the continuous relaxation space.
AQLM~\citep{egiazarian2024aqlm} employs additive multi-codebook quantization with joint optimization across blocks, achieving Pareto-optimal accuracy-versus-size trade-offs at 2 to 3 bits.

\paragraph{Extreme Low-bit Quantization.}
Pushing below 2 bits per weight requires moving beyond uniform formats.
OneBit~\citep{xu2024onebit} demonstrates that 1-bit factors can be stabilized through row and column scaling.
DBF~\citep{boza2026addition} combines two binary matrix multiplications with interleaved diagonal scalings for hardware-efficient binary inference.
LittleBit~\citep{lee2025littlebit} enhances extreme-bit accuracy via multi-scale scaling and residual compensation.
MDBF~\citep{ichikawa2025mdbf} identifies the single-envelope bottleneck in DBF and replaces the rank-one magnitude envelope with a rank-$l$ multi-envelope.
VPTQ~\citep{liu2024vptq} employs vector quantization using lattice codebooks.
BitNet~\citep{wang2023bitnet} explores 1-bit transformers trained from scratch.

\paragraph{Fine-Tuning for Quantized Models.}
QLoRA~\citep{dettmers2023qlora} freezes the quantized backbone and trains low-rank adapters, enabling fine-tuning on consumer GPUs.
\onecomp supports QLoRA as a final optional stage for both domain adaptation and quantization error recovery.

\section{Conclusions}
\label{sec:conclusions}

\onecomp provides a fully automated, resource-adaptive PTQ toolkit for foundation models.
Given a model identifier, it automatically constructs a compression plan, performs calibration and mixed-precision allocation, and executes the appropriate optimization pipeline within the available hardware budget.
 \onecomp transforms quantization from a fragmented collection of techniques into a cohesive, production-ready workflow that scales seamlessly from single-GPU layer-wise PTQ to more powerful block-wise and global refinement methods, while also supporting parameter-efficient fine-tuning.

At the algorithmic level, \onecomp supports a broad and practical range of precision settings.
\jointq significantly outperforms standard GPTQ at 3- and 4-bit precision by jointly optimizing integer assignments and quantization scales.
At the extreme low-bit frontier, DBF and \mdbf~\citep{ichikawa2025mdbf} preserve meaningful accuracy at 1--2 bits, while \autobit automates heterogeneous bit allocation under explicit resource constraints.
On the optimization side, \qep~\citep{arai2025quantization} bridges much of the gap between lightweight layer-wise PTQ methods and more expensive block-wise and global PTQ methods by explicitly modeling and compensating for error propagation from upstream quantized layers in closed form.
\lpcd~\citep{ichikawa2025lpcd} provides a unified framework for optimizing functionally coupled submodules.
Across the LLaMA and Qwen families, our experiments demonstrate that \onecomp consistently outperforms existing toolchains, delivers monotonic quality gains as GPU resources increase, and significantly benefits from larger calibration sets.

Looking ahead, our ambitions extend beyond quantization.
We aim to develop \onecomp into a comprehensive compression platform for foundation models that incorporates techniques such as activation quantization, KV-cache compression, knowledge distillation, structured pruning, and other compression methods within a unified, resource-adaptive interface.
Our immediate roadmap includes activation quantization and KV-cache compression, both of which are essential for efficient long-context inference and integrated as primary targets within \onecomp.
We also plan to release the \mdbf multi-envelope format and the diversity-aware calibration sampler in future releases.
Our long-term objective extends beyond improving compression performance; we aim to reshape the economics of foundation models.
\onecomp is guided by a vision for a future where progress relies less on brute-force GPU scaling and more on efficiency as a primary driver of innovation.
In this view, next-generation model systems should not only be trained and served with greater hardware resources but also be designed, compressed, and deployed much more efficiently.
Our goal is for \onecomp to become the standard open platform for facilitating and accelerating this transition.

\paragraph{Open-Source Vision.}
We release \onecomp as open-source software to help establish a shared foundation for the future of model compression.
By making \onecomp publicly available, we aim to accelerate innovation in research and engineering, support emerging architectures and hardware backends, and enable the broader community to extend the platform with new compression methods and capabilities.
We envision \onecomp evolving into a widely adopted open platform that defines the next generation of efficient AI systems.

\bibliographystyle{unsrtnat}
\bibliography{ref}

\newpage
\appendix
\section{Notation}
\label{sec:notation}

For convenience, Table~\ref{tab:notation} collects the main symbols that appear throughout this paper.
Symbols introduced with a local scope in a single equation are not repeated here; the reader is directed to the defining equation in each case.
Boldface lowercase letters (e.g.\ $\bm{\theta}$) denote vectors, uppercase letters (e.g.\ $W$) denote matrices, and calligraphic letters (e.g.\ $\mathcal{Q}$) denote sets.
A hat accent ($\widehat{\cdot}$) consistently marks a quantized or perturbed counterpart of the corresponding full-precision quantity.

\begin{table}[h]
\centering\small
\caption{Summary of principal notation.}
\label{tab:notation}
\begin{tabular}{cl}
\toprule
Symbol & Description \\
\midrule
\multicolumn{2}{l}{Model and data} \\
$L$ & Number of transformer blocks \\
$d,~ d_k,~ d_{\mathrm{ff}}$ & Hidden, per-head, and feed-forward dimensions \\
$H,~ H_{\mathrm{kv}}$ & Number of attention / key-value heads \\
$N,~ M,~ T$ & Weight input/output dimensions, sequence length \\
$P,~ n$ & Total parameters, number of calibration samples \\
$W_l,~\widehat{W}_l$ & Full-precision / quantized weight matrix of layer $l$ \\
$X_l,~\widehat{X}_l$ & Full-precision / perturbed input activations at layer $l$ \\
$Y_l = X_l W_l$ & Full-precision output at layer $l$ \\
$W_q, W_k, W_v, W_o$ & Attention projection matrices \\
$W_{\mathrm{gate}}, W_{\mathrm{up}}, W_{\mathrm{down}}$ & MLP projection matrices \\
$f_{\bm{\theta}},~ f_{\hat{\bm{\theta}}}$ & Full-precision / quantized model \\
$\mathcal{P},~\mathcal{C},~\mathcal{V}$ & Data distribution, calibration dataset, vocabulary \\
\midrule
\multicolumn{2}{l}{Quantization} \\
$b$ & Target bit-width \\
$s,~ z$ & Quantization scale and zero-point \\
$q,~ \widehat{w}$ & Quantized integer value, dequantized weight \\
$q_{\min},~ q_{\max}$ & Minimum / maximum of the integer range \\
$G,~ g(i,j)$ & Group size and group assignment function \\
$\mathcal{Q},~ Q$ & Feasible weight-matrix set, integer codebook \\
$\mathrm{cost}(l,c),~\mathrm{err}(l,c)$ & Memory cost / predicted error for config $c$ (\autobit) \\
\midrule
\multicolumn{2}{l}{Optimization} \\
$H_l = X_l^\top X_l$ & Input Gram matrix (Hessian approximation) \\
$\widehat{H}_l = \widehat{X}_l^\top \widehat{X}_l$ & Gram matrix from perturbed activations \\
$\bm{\delta}_l,~\bm{C}_l = \widehat{X}_l^\top \bm{\delta}_l$ & Activation error, error-propagation matrix (\qep) \\
$\alpha_l$ & \qep propagation-strength parameter \\
$\mathcal{J}$ & Submodule reconstruction objective (\lpcd) \\
$\Pi_{\mathcal{Q}}$ & Projection onto $\mathcal{Q}$ (any layer-wise quantizer) \\
$\lambda,~\tau,~k,~\rho$ & Regularization, KL temp., STE temp., SAM radius \\
$\mathcal{L}_{\mathrm{KL}},~\mathcal{L}_{\mathrm{inter}},~\mathcal{L}_{\mathrm{ent}}$ & KL / hidden-state alignment / entropy losses \\
$\bm{z}_j^{(\mathrm{fp})},~\bm{z}_j^{(\mathrm{q})}$ & Teacher / student logits at token position $j$ \\
\midrule
\multicolumn{2}{l}{DBF / MDBF} \\
$S_a,~ S_b \in \{\pm 1\}$ & Binary sign matrices \\
$D_{\bm{a}},~ D_{\bm{m}},~ D_{\bm{b}}$ & Diagonal scaling matrices \\
$A, Q, B, G$ & Low-rank envelope factors (MDBF) \\
$R$ & Factorization rank \\
\midrule
\multicolumn{2}{l}{Preprocessing} \\
$\bm{s}$ & Per-channel scaling vector (SmoothQuant) \\
$R \in \mathrm{O}(N)$ & Orthogonal rotation matrix \\
\midrule
\multicolumn{2}{l}{Operators} \\
$\lfloor \cdot \rceil,~\mathrm{clamp}$ & Rounding to nearest integer, clamping \\
$\odot,~\|\cdot\|_F$ & Element-wise product, Frobenius norm \\
$\mathrm{diag}(\cdot),~\mathrm{tr}(\cdot)$ & Diagonal matrix, trace \\
$\sigma(\cdot),~\mathrm{CosSim}(\cdot,\cdot)$ & Sigmoid function, cosine similarity \\
\bottomrule
\end{tabular}
\end{table}

\end{document}